\documentclass[journal,a4paper]{IEEEtran}
\pdfoutput=1 

\usepackage{cite}

\ifCLASSINFOpdf
  \usepackage{graphicx}
  \graphicspath{{figures}{portraits}}
\else
  \usepackage[dvips]{graphicx}
  \graphicspath{{figures}{portraits}}
  \DeclareGraphicsExtensions{.eps,.ps}
\fi
\usepackage[cmex10]{amsmath}
\usepackage{array}
\usepackage{url}

\usepackage{color}
\usepackage{import} 
\usepackage{bm} 
\usepackage{bigints} 
\usepackage{amssymb, amsfonts, mathtools}
\usepackage{cases}

\usepackage[noabbrev]{cleveref} 
\crefformat{equation}{(#1)} 
\Crefformat{equation}{(#1)} 
\let\origref\ref 
\let\ref\Cref    

\usepackage{pgfplots}
\usepackage{tikz}
\usepgfplotslibrary{groupplots}
\usetikzlibrary{pgfplots.groupplots}
\pgfplotsset{compat=newest}
\pgfplotsset{plot coordinates/math parser=false}

\usepackage{nohyperref}

\newcommand{\ndot}[1]{\ensuremath{\protect\dot{#1}}} 

\newcommand{\abs}[1]{\ensuremath{\left\vert #1 \right\vert}}
\newcommand{\card}[1]{\ensuremath{\left\vert #1 \right\vert}}
\newcommand{\norm}[2]{\ensuremath{\left\Vert #1 \right\Vert_{#2}}}

\newcommand{\ud}{\ensuremath{\,\mathrm{d}}} 
\newcommand{\ppart}[1]{\ensuremath{\left[#1\right]^{+}}}

\newcommand{\R}{\ensuremath{\mathbb{R}}}  
\newcommand{\Rnn}{\ensuremath{\R_{\geq0}}} 
\newcommand{\Rp}{\ensuremath{\R_{>0}}} 

\DeclareMathOperator*{\mini}{minimize}

\newcommand{\minimizex}[1]{\ensuremath{& \text{$\mini_{#1}$} & &}}

\newcommand{\st}{\ensuremath{& \text{$\operatorname{subject~to}$} & &}}
\newcommand{\stff}{\ensuremath{&& &}}

\renewcommand{\vec}[1]{\ensuremath{\bm{\MakeLowercase{#1}}}}
\newcommand{\mat}[1]{\ensuremath{\bm{\MakeUppercase{#1}}}}
\newcommand{\vecidx}[2]{\ensuremath{\MakeLowercase{#1}_{#2}}}
\newcommand{\vecidxB}[2]{\ensuremath{\left[#1\right]_{#2}}}
\newcommand{\vecidxBppart}[2]{\ensuremath{\left[#1\right]^{+}_{#2}}}
\newcommand{\vecidxBnpart}[2]{\ensuremath{\left[#1\right]^{-}_{#2}}}
\newcommand{\matidx}[3]{\ensuremath{\MakeLowercase{#1}_{#2 #3}}}
\newcommand{\matidxB}[3]{\ensuremath{\left[#1\right]_{#2 #3}}}

\newcommand{\vecselect}[2]{\ensuremath{\left[#1\right]_{#2}}} 
\newcommand{\matcselect}[2]{\ensuremath{\left[#1\right]_{: #2}}} 
\newcommand{\transp}[1]{\ensuremath{#1^{\scriptscriptstyle\bm T}}}

\newcommand{\opt}[1]{\ensuremath{#1^{\star}}}
\newcommand{\basisvec}{\vec{e}}
\newcommand{\imat}{\mat{i}}
\newcommand{\zerovec}{\vec{0}} 

\newcommand{\onevec}{\vec{1}} 

\newcommand{\expop}[1]{\ensuremath{\mathrm{E}\left[#1\right]}}
\newcommand{\varop}[1]{\ensuremath{\mathrm{Var}\left[#1\right]}}
\newcommand{\covop}[1]{\ensuremath{\mathrm{Cov}\left[#1\right]}}
\newcommand{\traceop}[1]{\ensuremath{\mathrm{Tr}\left[#1\right]}}

\newcommand{\dB}{\text{dB}}

\newcommand{\lsmat}{a} 
\newcommand{\lsvar}{x} 
\newcommand{\lsin}{b} 
\newcommand{\lsobj}{f_0} 
\newcommand{\lsineq}{f} 
\newcommand{\lslagmul}{\lambda} 
\newcommand{\lsmatB}{p} 
\newcommand{\lsinB}{q}
\newcommand{\lslowbox}{l}
\newcommand{\lshighbox}{h}

\newcommand{\discnnint}{\ensuremath{\int^{+}}} 
\newcommand{\discindim}{M}
\newcommand{\discoutdim}{N}
\newcommand{\discstatedim}{\discoutdim}
\newcommand{\discin}{u} 
\newcommand{\discstate}{x} 
\newcommand{\discintc}{c} 
\newcommand{\discout}{y} 
\newcommand{\discmat}{\lsmat}
\newcommand{\discintin}{\protect\tilde\discstate}
\newcommand{\discint}{\discnnint} 
\newcommand{\discfun}{f_x} 
\newcommand{\discoutfun}{f_y} 
\newcommand{\discequi}{e} 
\newcommand{\discI}{\ensuremath{\mathcal{I}}} 
\newcommand{\discIpos}{\discI^{+}} 
\newcommand{\discIzero}{\discI^{0}} 
\newcommand{\discIneg}{\discI^{-}} 
\newcommand{\disclagmul}{\lambda} 
\newcommand{\disclowbox}{\lslowbox} 
\newcommand{\dischighbox}{\lshighbox} 
\newcommand{\discttset}{\mathcal{T}} 
\newcommand{\discconstin}{\lsin}
\newcommand{\discconvtime}{T}

\newcommand{\discintinB}{\protect\hat\discstate}

\newcommand{\lyap}{V}

\newcommand{\tdiscin}{\protect\hat\discin}
\newcommand{\tdiscstate}{z}
\newcommand{\tdiscfun}{f_z}

\newcommand{\sysin}{\discin}
\newcommand{\sysout}{\discout}
\newcommand{\sysstate}{\discstate}
\newcommand{\sysequi}{\discequi}
\newcommand{\sysstart}{0}
\newcommand{\sysindim}{\discindim}
\newcommand{\sysoutdim}{\discoutdim}
\newcommand{\sysstatedim}{R}
\newcommand{\sysfun}{\discfun}
\newcommand{\sysoutfun}{\discoutfun}
\newcommand{\sysconst}{c}

\newcommand{\olfmat}{w}
\newcommand{\olfmcout}{x}
\newcommand{\olfmcin}{\protect\tilde\olfmcout}
\newcommand{\olfodour}{o}
\newcommand{\olfin}{u}

\newcommand{\olfglout}{g}
\newcommand{\olfgc}{l}
\newcommand{\olfnumGL}{N}
\newcommand{\olfmatrowdim}{\csmeasdim}

\newcommand{\olfnnlsopt}{\olfmcout^\star}
\newcommand{\olftrue}{0}
\newcommand{\olfnoise}{\eta}

\newcommand{\matmean}{\mu_{\olfmat}}
\newcommand{\matvar}{\sigma^2_{\olfmat}}
\newcommand{\truemixmean}{\mu_{\olfmcout}}
\newcommand{\truemixvar}{\sigma^2_{\olfmcout}}

\newcommand{\noisevar}{\sigma^2_{\olfnoise}}
\newcommand{\noisestd}{\sigma_{\olfnoise}}

\newcommand{\sparsity}{s}
\newcommand{\supp}{\ensuremath{\mathcal{S}}}
\newcommand{\compsupp}{\ensuremath{\mathcal{S}^\text{c}}}

\newcommand{\cssignal}{x}
\newcommand{\cstrue}{0}
\newcommand{\csmat}{A}
\newcommand{\csmeas}{b}
\newcommand{\csmeaserr}{\varepsilon}
\newcommand{\csmeasdim}{M}
\newcommand{\cssignaldim}{N}
\newcommand{\csreg}{\alpha}
\newcommand{\cserrconst}{c_{\csmeaserr}}
\newcommand{\cssrpconst}{c_0}
\newcommand{\cslincom}{h}
\newcommand{\csobjfun}{f}

\newcommand{\idxvar}{j}

\newcommand{\dummy}{\gamma}
\newcommand{\dummyB}{\theta}

\newcommand{\neighbourhood}{U}

\newcommand{\figpath}{figures/}
\newlength\figureheight
\newlength\figurewidth

\begin{document}
%
\title{A Discontinuous Neural Network\\for Non-Negative Sparse Approximation}
%
%
%

\author{Martijn~Arts,
        Marius~Cordts,
        Monika Gorin,
        Marc Spehr,
        and~Rudolf~Mathar \vspace{-0.5cm}
\thanks{M. Arts and R. Mathar are with the Institute for
        Theoretical Information Technology, RWTH Aachen University, Aachen,
        Germany.}
\thanks{M. Cordts is now with the Environment Perception Dept. at Daimler R\&D, Sindelfingen, Germany.}
\thanks{M. Gorin and M. Spehr are with the Department of Chemosensation, Institute for Biology II, RWTH Aachen University, Aachen, Germany.}
\thanks{This work was partly supported by the Deutsche For\-schungs\-ge\-mein\-schaft (DFG) SPP 1395: Informations- und Kommunikations\-theorie in der
Molekularbiologie (InKoMBio), Grants: MA 1184/20-2 \& SP 724/8-1.}
\thanks{Please direct your e-mail correspondence to: arts@ti.rwth-aachen.de}
\thanks{This work has been submitted to the IEEE for possible publication.
	Copyright may be transferred without notice, after which this version may
	no longer be accessible.}}

\maketitle

\begin{abstract}
This paper investigates a discontinuous neural network which is used as a model of the mammalian olfactory system and can more generally be applied to solve non-negative sparse approximation problems. By inherently limiting the systems integrators to having non-negative outputs, the system function becomes discontinuous since the integrators switch between being inactive and being active. It is shown that the presented network converges to equilibrium points which are solutions to general non-negative least squares optimization problems. We specify a Caratheodory solution and prove that the network is stable, provided that the system matrix has full column-rank. Under a mild condition on the equilibrium point, we show that the network converges to its equilibrium within a finite number of switches. Two applications of the neural network are shown. Firstly, we apply the network as a model of the olfactory system and show that in principle it may be capable of performing complex sparse signal recovery tasks. Secondly, we generalize the application to include  non-negative sparse approximation problems and compare the recovery performance to a classical non-negative basis pursuit denoising algorithm. We conclude that the recovery performance differs only marginally from the classical algorithm, while the neural network has the advantage that no performance critical regularization parameter has to be chosen prior to recovery.
\end{abstract}


%
\IEEEpeerreviewmaketitle


%
%
%
%
\section{Introduction}
\label{sec:introduction}

\IEEEPARstart{S}{parsity} is a concept that has recently generated a considerable amount of interest in mathematics, computer science and electrical engineering. In essence, it states that many signals may be represented by only a small number of non-zero coefficients using a suitable basis or dictionary. 
The field of compressed sensing exploits knowledge about the sparsity which allows to recover signals from a smaller number of measurements compared to classical methods \cite{donoho_compressed_2006,candes_robust_2006}. This is done by solving optimization problems that penalize non-sparse solution candidates. 

In this work, we study a sparse signal recovery problem faced by the mammalian main olfactory system, namely the decomposition of a mixture of chemical stimuli from noisy measurements. The result of our modeling is a discontinuous neural network that solves a quadratic program (QP) in order to recover the signal of interest from the measurements. We later show that this network can be applied more generally to solve non-negative sparse approximation problems, which do not have to be related to the biological application.

The connection between neural networks and optimization problems has been investigated for several decades now. A particularly interesting class, called (continuous) \emph{Hopfield Neural Networks} (HNNs), was proposed by Hopfield in the 1980s \cite{hopfield_neural_1982,hopfield_neural_1985}. These networks consider either weighted excitatory or inhibitory feedback between several ``neurons''. The neurons themselves are modeled as amplifiers with a potentially arbitrary input-output function. A sufficient condition for stability of such a network is that the synaptic weights are symmetric and feedback loops, i.e., the output of a neuron is connected to its own input, are not present. Such networks can be used to solve a variety of optimization problems, see \cite{wen_review_2009} for a review.
In the context of QPs, known networks used to solve (bounded) QPs such as \cite{bouzerdoum_neural_1993,xia_new_1996} are functionally related to the network presented in this work. These networks, however, differ fundamentally in structure and use other methods to investigate the convergence behavior compared to our approach.
Specifically for sparse approximation scenarios, Locally Competitive Algorithms (LCAs) \cite{rozell_sparse_2008} were proposed. They solve optimization problems with a quadratic objective function, regularized by a cost function that is applied to the output of the integrators. Depending on the choice of the cost function, these neural networks solve the well known basis pursuit denoising problem (see also \ref{eq:bpdn}) or an approximation of the original non-convex compressed sensing problem, where each non-zero entry is penalized independent of its actual amplitude (cf. \ref{eq:l0minimization}). The convergence behavior of the LCAs was studied in detail in \cite{balavoine_convergence_2012}.

The contributions of this paper are summarized in the following. We study a neural network in the form of a dynamic system where the integrators of the system are limited to having non-negative outputs. This results in discontinuities of the system function, since the integrators switch between being active and inactive depending on the value of their output. First, we show that the neural networks equilibrium points are the minimizers of a general non-negative least squares optimization problem, then we specify a Caratheodory solution for the systems discontinuous differential equations. Having defined the Caratheodory solution, we use Lyapunov's direct method to show that the neural network is asymptotically stable, provided that the system matrix has full column-rank. Under a mild condition on the equilibrium point, we show that the network converges to this point within a finite number of switches. Subsequently, we discuss that the non-negative integrators can be generalized to arbitrary lower and upper limits, so that the equilibrium point corresponds to the solution of a bounded least squares problem. Two different applications of the neural network are presented. The first is the original application, which served as the foundation for the presented neural network. It is derived as a model of the mammalian olfactory system and we investigate the performance of a stimulus mixture decomposition scenario. Based on the network performance, we conclude that the olfactory system may indeed recover the composition of chemical stimuli from a noisy mixture of neural responses. The second application deals with the more general class of non-negative sparse approximation problems. Here, we compare the recovery performance of the neural network with a non-negative basis pursuit denoising method and show that for this type of application the stability of the network remains unaffected even if the system matrix does not have full column-rank. From this investigation, we conclude that the neural network performance differs only marginally compared to the ``classical'' recovery method. However, the network has the additional advantage that no performance critical regularization parameter has to be determined prior to recovery.

An overview of the structure of this paper is given in the following. In \ref{sec:cssa}, we give the necessary background about compressed sensing / sparse approximation and summarize relevant results on the when dealing with non-negative sparse recovery problems. A concise overview of the stability theory of (discontinuous) dynamic systems can be found in \ref{sec:disc_dyn_sys}. The discontinuous neural network which is considered in this work is presented in \ref{sec:system_model} and relevant notation is defined. A formal investigation of the convergence properties and stability analysis of the neural network is given in \ref{sec:conv_analysis}. We present two applications of the presented neural network. Firstly, it is used as a model of the mammalian olfactory system, where it is shown that such a network would be able to decompose a mixture of chemical stimuli (\ref{sec:olf_app}). Secondly, in \ref{sec:nnsa_app}, we show that the network is more generally able to solve non-negative sparse approximation problems. \ref{sec:conclusion} concludes the paper.


\section{Background}
\subsection{Compressed Sensing / Sparse Approximation}
\label{sec:cssa}
Compressed sensing is a relatively new research field that deals with recovering signals from a lower number of linear measurements compared to classical methods of signal recovery \cite{donoho_compressed_2006,candes_robust_2006}. The field relies on a concept called \emph{sparsity}ſ, i.e., the assumption that the signal can be represented by only a few nonzero coefficients in a suitable basis. A signal $\vec{\cssignal} \in \R^\cssignaldim$ is called $\sparsity$-sparse if at most $\sparsity$ coefficients of the vector $\vec{\cssignal}$ are nonzero, that is, if $\norm{\vec{\cssignal}}{0} \leq \sparsity$. Here, $\norm{\vec{\cssignal}}{0} := \#\{\idxvar : \vecidx{\cssignal}{\idxvar} \neq 0\}$ denotes the $\ell_0$ pseudo-norm. Note that this "norm" does not fulfill the homogeneity property and therefore does not represent a proper norm.

The task in compressed sensing is to recover an $\sparsity$-sparse signal $\vec{\cssignal}$ from a measured signal $\vec{\csmeas} \in \R^\csmeasdim$, obtained by a linear measurement process described by $\vec{\csmeas} = \mat{\csmat}\vec{\cssignal}$ with $\mat{\csmat} \in \R^{\csmeasdim\times\cssignaldim}$. Here, the number of measurements is small compared to the dimension of the original signal, i.e., $\csmeasdim \ll \cssignaldim$. Ideally, one would like to recover $\vec{\cssignal}$ using the following optimization problem:
\begin{equation}
    \label{eq:l0minimization}
    \begin{aligned}
        \minimizex{\vec{\cssignal}} \norm{\vec{\cssignal}}{0} \\
        \st \vec{\csmeas} = \mat{\csmat}\vec{\cssignal}.
    \end{aligned}
\end{equation}
However, solving problem \ref{eq:l0minimization} is NP-hard and thus intractable \cite{candes_robust_2006}. If the matrix $\mat{\csmat}$ fulfills certain conditions however, e.g., the so-called null space property or the restricted isometry property, the minimizer of the $\ell_1$-minimization problem (also called basis pursuit)
\begin{equation}
    \label{eq:l1minimization}
    \begin{aligned}
        \minimizex{\vec{\cssignal}} \norm{\vec{\cssignal}}{1} \\
        \st \vec{\csmeas} = \mat{\csmat}\vec{\cssignal},
    \end{aligned}
\end{equation}
is also the minimizer of problem \ref{eq:l0minimization} and hence may recover the desired signal vector exactly \cite{donoho_compressed_2006,candes_robust_2006}. In problem \ref{eq:l1minimization}, the $\ell_1$-norm, which is the convex relaxation of the $\ell_0$-``norm'', was used as the objective. Note, that we confined $\vec{\cssignal}$, $\mat{\csmat}$ and $\vec{\csmeas}$ to the real numbers but the results summarized above also hold in the complex setting.

In practical scenarios, virtually all measurements are perturbed by some form of noise. Thus, in practice the measurement process may be modeled as $\vec{\csmeas} = \mat{\csmat}\vec{\cssignal} + \vec{\csmeaserr}$ where $\vec{\csmeaserr} \in \R^\csmeasdim$ denotes the noise component. In this case, taking $\csmeasdim=\cssignaldim$ measurements will not guarantee exact recovery of the desired signal vector anymore. Hence, one would typically take more measurements ($\csmeasdim>\cssignaldim$) and try to recover a good approximation of the signal vector through least-squares minimization. Likewise, in compressed sensing, one cannot hope to recover the exact signal vector when measurements are noisy. Consequently, the linear constraint $\vec{\csmeas} = \mat{\csmat}\vec{\cssignal}$ in problem \ref{eq:l1minimization} is too strict since it does not allow for deviations caused by noise. To cope with this, one may replace the constraint in problem \ref{eq:l1minimization} by a quadratic constraint $\norm{\mat{\csmat}\vec{\cssignal} - \vec{\csmeas}}{2}^2 \leq \cserrconst$ which allows for deviations within a sphere with constant radius. To choose $\cserrconst$ meaningfully, it is necessary to have prior knowledge about the noise power. Another method is regularizing the objective function, as done in basis pursuit denoising (BPDN):
\begin{equation}
    \label{eq:bpdn}
    \begin{aligned}
        \minimizex{\vec{\cssignal}} \frac{1}{2}\norm{\mat{\csmat}\vec{\cssignal} - \vec{\csmeas}}{2}^2 + \csreg \norm{\vec{\cssignal}}{1}.
    \end{aligned}
\end{equation}
Here, $\csreg$ is the regularization parameter which allows to tune the influence of the two terms in the objective. Naturally, exact recovery of the signal vector cannot be guaranteed in the presence of noise, however, under certain conditions the recovery error can be bounded \cite{candes_stable_2006} w.r.t. $\cserrconst$.

The discovery of recovery guarantees and error bounds resulted in compressed sensing becoming a very active research field. Many applications have been investigated, among them magnetic resonance imaging (MRI), data acquisition and a multitude of signal processing tasks \cite{candes_introduction_2008}. A related research field is sparse approximation. While in compressed sensing it is typically assumed that the matrix $\mat{\csmat}$ may be modified, in sparse approximation the matrix is fixed and given by the application. Since the recovery algorithms are essentially the same, we do not make a special distinction between these two fields here.

Quite surprising results have been reported when the recovery of non-negative signal vectors, i.e., $\vec{\cssignal} \in \Rnn^\cssignaldim$ is desired. Such non-negativity constraints appear naturally in, e.g., image processing, text \& data mining and environmetrics (see also \cite{chen_nonnegativity_2009}). In the noiseless case, the non-negative signal recovery problem can be formalized as:
\begin{equation}
    \label{eq:nnl0minimization}
    \begin{aligned}
        \minimizex{\vec{\cssignal}} \norm{\vec{\cssignal}}{0} \\
        \st \vec{\csmeas} = \mat{\csmat}\vec{\cssignal} \\
        \stff \vec{\cssignal} \geq \zerovec.
    \end{aligned}
\end{equation}
Interestingly, the solution space $\{ \vec{\cssignal}~|~\mat{\csmat}\vec{\cssignal} = \vec{\csmeas}, \vec{\cssignal} \geq \zerovec \}$ is a singleton, provided that certain conditions on $\vec{\cssignal}$ and the matrix $\mat{\csmat}$ are fulfilled \cite{bruckstein_uniqueness_2008,wang_conditions_2009}. Firstly, $\vec{\cssignal}$ must be sufficiently sparse, where a threshold on the sparsity is defined by the one-sided coherence of the matrix $\mat{\csmat}$. The condition on the matrix states that the row-span of $\mat{\csmat}$ must intersect the positive orthant, i.e., a linear combination $\vec{\cslincom}$ must exist, such that $\transp{\vec{\cslincom}}\mat{\csmat} > 0$. 

Let us consider an optimization problem with a general objective function $\csobjfun$ as well as constraints $\mat{\csmat}\vec{\cssignal} = \vec{\csmeas}$ and $\vec{\cssignal} \geq \zerovec$. Furthermore, let the conditions discussed above be fulfilled. Then, since the solution to the linear equation system with non-negativity constraint is unique, for a broad class of objective functions $\csobjfun$, including the sparsity inducing $\ell_0$-``norm'' and $\ell_1$-norm, it is the solution to the optimization problem as well \cite{bruckstein_uniqueness_2008}. This means, that problem \ref{eq:nnl0minimization} can be solved efficiently by a linear program, a result which is similar to the classical compressed sensing scenario.

Turning to the noisy case again, where the measurements are obtained by $\vec{\csmeas} = \mat{\csmat}\vec{\cssignal} + \vec{\csmeaserr}$, perfect recovery is generally not possible. Similar to the classical compressed sensing case, one could add the non-negativity constraint to problem \ref{eq:bpdn}, resulting in the non-negative basis pursuit denoising (NNBPDN) problem. Inspired by the noiseless results above, one could argue that the sparsity inducing objective function may be redundant, so that solving a non-negative least squares (NNLS) problem (see problem \ref{eq:nnls}) would suffice. This idea was investigated in \cite{slawski_non-negative_2013}, where subsequent thresholding was suggested to gain truly sparse solutions. In the same work, error bounds were given under the condition that the matrix $\mat{\csmat}$ fulfills the self-regularizing property: $\transp{\vec{\cslincom}}(\frac{1}{\csmeasdim}\transp{\mat{\csmat}}\mat{\csmat}) \vec{\cslincom} \geq \cssrpconst^2 (\transp{\onevec} \vec{\cslincom})^2  \quad \forall \vec{\cslincom} \geq 0$, where $\cssrpconst \geq 0$ is a universal constant. One advantage of the NNLS approach to solving non-negative sparse approximation problems is that the method is independent of any regularization parameters.

\subsection{(Discontinuous) Dynamical Systems}
\label{sec:disc_dyn_sys}
Dynamical systems with multiple-inputs and multiple-outputs are commonly described in state space. Defining the input vector $\vec{\sysin}(t) \in \R^\sysindim$, the output vector $\vec{\sysout}(t) \in \R^\sysoutdim$ and the state vector $\vec{\sysstate}(t) \in \R^\sysstatedim$, the dynamical system can be described as
\begin{equation}
\begin{aligned}
\ndot{\vec{\sysstate}}(t) &= \vec{\sysfun}(\vec{\sysstate}(t),\vec{\sysin}(t)), \\
\vec{\sysout}(t) &= \vec{\sysoutfun}(\vec{\sysstate}(t),\vec{\sysin}(t)).
\end{aligned}
\label{eq:mimosys}
\end{equation}
Here, $\vec{\sysfun}: \R^{\sysstatedim} \times \R^{\sysindim} \to \R^\sysstatedim$ is called the system function and $\vec{\sysoutfun}: \R^{\sysstatedim} \times \R^{\sysindim} \to \R^\sysoutdim$ is called the output function. The term $\ndot{\vec{\sysstate}}(t)$ is the time derivative of the states, i.e., $\ndot{\vec{\sysstate}}(t) = \ud \vec{\sysstate}(t) / \ud t$.

In many cases, it is desirable to analyze dynamical systems for stability. Stable systems ensure a certain level of predictability, thus this concept is very important in control theory. We briefly revise the definitions of stability, which are relevant to this work, following \cite{adamy_grundlagen_2009}. A point $\vec{\sysstate}_\sysequi$ of a dynamical system with static input $\vec{\sysin}(t) = \vec{\sysconst}_\sysin$ is called \emph{equilibrium point} if
\begin{equation*}
\ndot{\vec{\sysstate}}(t) = \vec{\sysfun}(\vec{\sysstate}_\sysequi, \vec{\sysin}(t) = \vec{\sysconst}_\sysin) = \zerovec.
\end{equation*}
We assume $\vec{\sysstate}_\sysequi = \zerovec$ and $\vec{\sysconst}_\sysin = \zerovec$ without loss of generality in the following since every system can be transformed to this case by a simple linear transformation. An equilibrium point $\vec{\sysstate}_\sysequi$ is called (locally) attractive if all starting points within a neighborhood around the origin $\vec{\sysstate}_\sysstart \in \neighbourhood(\zerovec)$ cause a trajectory $\lim_{t \to \infty} \vec{\sysstate}(t) = \vec{\sysstate}_\sysequi = \zerovec$ for the static input $\vec{\sysin}(t) = \zerovec$. If this neighborhood is $\neighbourhood(\zerovec) = \R^\sysstatedim$, $\vec{\sysstate}_\sysequi$ is called globally attractive. Additionally, we call an equilibrium point Lyapunov stable if for every $\varepsilon$-neighborhood $\neighbourhood_\varepsilon(\zerovec) = \{\vec{\dummy} \in  \R^\sysstatedim~|~\norm{\vec{\dummy}}{2} < \varepsilon\}$, there is a $\delta$-neighborhood $\neighbourhood_\delta(\zerovec) = \{\vec{\dummy} \in  \R^\sysstatedim~|~\norm{\vec{\dummy}}{2} < \delta\}$ such that
if $\vec{\sysstate}_0 \in \neighbourhood_\delta(\zerovec)$, then $\vec{\sysstate}(t) \in \neighbourhood_\varepsilon(\zerovec)$ for every $t > 0$ and $\delta = \delta(\varepsilon)$. Finally, if the equilibrium is locally (globally) attractive and Lyapunov stable, it is called locally (globally) asymptotically stable.
In summary, if asymptotic stability holds, the dynamical system with constant input and a starting point $\vec{\sysstate}_0$ close enough to the equilibrium point will generate a trajectory converging to $\vec{\sysstate}_\sysequi$ while the trajectory is guaranteed to stay within a certain range of $\vec{\sysstate}_\sysequi$ in the process.

For \emph{classical} dynamical systems, when analyzing stability there is the restriction that the system function $\vec{\sysfun}$ must be continuously differentiable \cite{slotine_applied_1991} or more generally Lipschitz continuous \cite{khalil_nonlinear_2002,sastry_nonlinear_1999}. This guarantees that a solution $\vec{\sysstate}(t)$ to the differential vector equation \ref{eq:mimosys} exists and that it is unique for a starting point $\vec{\sysstate}_0$. When studying \emph{discontinuous} dynamical systems, this restriction is dropped so that $\vec{\sysfun}$ can be discontinuous. Stability analysis of such systems is more complicated, since existence and uniqueness of solutions is not guaranteed anymore. Also, there are different kinds of generalized solutions which have been studied, such as Caratheodory or Filippov solutions \cite{cortes_discontinuous_2008}. In this work, we make use of a Caratheodory solution. In simple terms, such a solution generally satisfies the system equations, except for a set of time instances of measure zero (e.g., a countable, not necessarily finite, set). Formally, a Caratheodory solution can be defined using the Lebesgue integral as an absolutely continuous function $\vec{\sysstate}(t)$, which satisfies \cite{cortes_discontinuous_2008}
\begin{equation*}
\vec{\sysstate}(t) = \vec{\sysstate}_0 + \int_{0}^{t} \vec{\sysfun}(\vec{\sysstate}(t'), \vec{\sysin}(t')) \ud t'.
\end{equation*}

\section{System Model}
\label{sec:system_model}
\begin{figure}[ht]
  \centering
  \def\svgwidth{0.75 \columnwidth}
  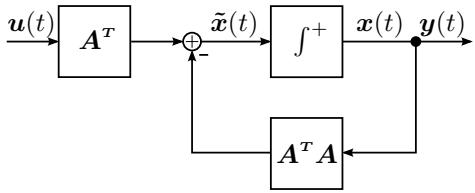
  \caption{Block diagram of the discontinuous system with a non-linear integrator.}
  \label{fig:disc_sys}
\end{figure}
\begin{figure}[!ht]
  \centering
  \import{figures/}{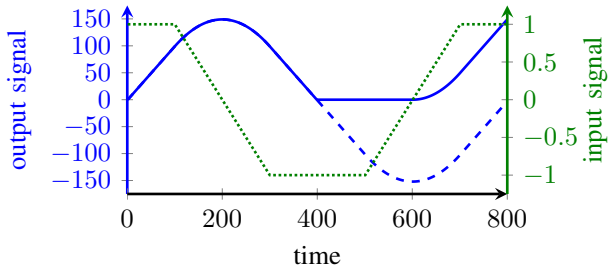}
  \caption{Comparison between the output signals of a standard linear integrator (dashed blue line) and a non-linear integrator as described above (solid blue line). The input signal (dotted green line) is plotted according to the second ordinate on the right.}
  \label{fig:lim_int_plot}
\end{figure}

We consider a continuously valued neural network with input vector $\vec{\discin}(t) \in \R^{\discindim}$ and non-negative output vector $\vec{\discout}(t) \in \Rnn^{\discoutdim}$. For the upcoming analysis, we represent the network as a discontinuous system in state space with state vector $\vec{\discstate}(t)$ (see also \ref{fig:disc_sys}). From this, it follows that the matrix $\mat{\discmat} \in \R^{\discindim \times \discoutdim}$. The integrator denoted by $\discnnint$ is a non-linear integrator as defined in the following. If all states are positive, this block behaves like a standard integrator, however, if a state reaches zero, integration of this state is halted. Thus, said state remains at value zero until it is subjected to a positive gradient again. Defining the non-linear integrator as $\vec{\discstate}(t) = \int_{0}^{t} \ndot{\vec{\discstate}}(\tau) \ud \tau$, it follows that the derivative of the state vector $\ndot{\vec{\discstate}}(t)$ must be discontinuous:
\begin{equation}
\begin{aligned}
\vecidx{\ndot{\discstate}}{i}(t) &= 
\begin{dcases}
\vecidx{\discintin}{i}(t), & \text{if } \vecidx{\discstate}{i}(t) > 0 \\
\ppart{\vecidx{\discintin}{i}(t)}\!\!\!\!, & \text{if } \vecidx{\discstate}{i}(t) = 0 \\
\discintc, & \text{if } \vecidx{\discstate}{i}(t) < 0.
\end{dcases} \quad i=1, \ldots, \discoutdim. \\
\end{aligned}
\label{eq:nnint}
\end{equation}
Here, $\vecidx{\discintin}{i}(t)$ is the input of the $i$-th integrator and the positive part of the $i$-th vector element is denoted as $\ppart{\vecidx{\discintin}{i}} = \max\{0, \vecidx{\discintin}{i}\}$. The inclusion of the constant $\discintc > 0$ is a technicality to ensure recovery from non-negative states due to faulty initial conditions or disturbances. See \ref{fig:lim_int_plot} for an example of
the behavior of the integrator described in \ref{eq:nnint}.

Using \ref{eq:nnint} and 
\begin{equation*}
\vec{\discintin}(t) = \transp{\mat{\discmat}} \vec{\discin}(t) - \transp{\mat{\discmat}}\mat{\discmat}\vec{\discstate}(t),
\end{equation*}
the system equations can be formalized for the discontinuous system depicted in \ref{fig:disc_sys}.
\begin{align}
\label{eq:disc_sys}
& \ndot{\vec{\discstate}}(t) \!\!\!\!\!\!\!\! & &= \; \vec{\discfun}(\vec{\discstate}(t),\vec{\discin}(t)), 
\;\; \text{with} \;\; \vecidx{\ndot{\discstate}}{i}(t) = \vecidx{\discfun}{i}(\vec{\discstate}(t),\vec{\discin}(t)), \nonumber \\
& \vecidx{\ndot{\discstate}}{i}(t) \!\!\!\!\!\!\!\! & &= \; 
\begin{dcases}
\vecidxB{\vec{\discintin}(t)}{i}, & \text{if } \vecidx{\discstate}{i}(t) > 0 \\
\vecidxBppart{\vec{\discintin}(t)}{i}\!\!, & \text{if } \vecidx{\discstate}{i}(t) = 0 \\
\multicolumn{1}{c}{$\discintc,$} & \text{if } \vecidx{\discstate}{i}(t) < 0
\end{dcases} \qquad i=1,\ldots,\discoutdim \\
& \vec{\discout}(t) \!\!\!\!\!\!\!\! & &= \; \vec{\discoutfun}(\vec{\discstate}(t),\vec{\discin}(t)) = \vec{\discstate}(t). \nonumber
\end{align}

The system defined by \ref{eq:disc_sys} has a very interesting property: it is able to solve non-negative least squares (NNLS) optimization problems. To further clarify this feature, consider the following optimization problem with optimal point $\opt{\vec{\lsvar}}$ and $\vec{\lsin} \in \R^{\discindim}$:
\pagebreak
\begin{equation}
\begin{aligned}
\minimizex{\vec{\lsvar}} \frac{1}{2} \norm{\mat{\lsmat}\vec{\lsvar}-\vec{\lsin}}{2}^{2} \\
\st \vec{\lsvar} \geq \zerovec.
\end{aligned}
\label{eq:nnls}
\end{equation}
This is a convex optimization problem, which does not possess an analytical solution, in contrast to an unconstrained least-squares problem \cite{boyd_convex_2004}. 

If the discontinuous system of \ref{fig:disc_sys} receives the constant input $\vec{\discin}(t) = \vec{\lsin}$, it will converge to its equilibrium point $\vec{\discstate_{\discequi}} = \opt{\vec{\lsvar}}$, which is also the optimal point of the NNLS optimization problem \ref{eq:nnls}. We formally prove this fact in \ref{sec:equi_point}.

\section{Convergence Analysis}
\label{sec:conv_analysis}
In this section, we take a closer look at the convergence behavior of the discontinuous dynamic system shown in \ref{fig:disc_sys}. First, we show that the equilibrium points of the system are solutions of NNLS optimization problems in \ref{sec:equi_point}. Second, we characterize the Caratheodory solution of the differential equation in \ref{sec:caratheo_solution}, which is a prerequisite for the stability analysis. Then, we turn to the actual stability analysis in \ref{sec:stability}, which is based on Lyapunov's direct method. Later, we additionally prove in \ref{sec:finite_switches} that the number of switches performed by the non-linear integrator during convergence to the equilibrium point is finite under a mild condition on the solution. Finally, we briefly discuss in \ref{sec:box_constraints} how the convergence analysis can be generalized to limited integrators with box constraints of the form $\vec{\disclowbox} \leq \vec{\discstate}(t) \leq \vec{\dischighbox}$. 

Before we begin the convergence analysis, we introduce some notation which will be helpful in subsequent sections of this text. 
First, we denote the set of all states' indices as $\discI = \{1,\dots, \discoutdim\}$. Then, we divide this set into three time-varying sets $\discIpos(t)$, $\discIzero(t)$ and $\discIneg(t)$. These  can informally be described as the indices of the system states that are positive, zero and negative at time $t$, respectively. A formal definition of the sets can be given as follows:
\begin{equation}
\begin{aligned}
& \discIpos(t) \!\!\!\!\! & &= \; \Big\{i\in\discI \,\big|\, \vecidx{\discstate}{i}(t) > 0 \Big\} \\
&              & & \quad \;\; \cup \Big\{i\in\discI \,\big|\, \vecidx{\discstate}{i}(t) = 0,\ \vecidx{\discintin}{i}(t) \geq 0 \Big\}, \\
& \discIzero(t) \!\!\!\!\! & &= \; \Big\{i\in\discI \,\big|\, \vecidx{\discstate}{i}(t) = 0,\ \vecidx{\discintin}{i}(t) < 0 \Big\}, \\
& \discIneg(t) \!\!\!\!\! & &= \; \Big\{i\in\discI \,\big|\, \vecidx{\discstate}{i}(t) < 0\Big\}.
\end{aligned}
\label{eq:disc_idx_sets}
\end{equation}
Note, that the set $\discIpos(t)$ also contains indices of states which are zero at time $t$ if its corresponding integrator input is positive. Also, no state index can be present in two or all of the sets $\discIpos(t)$,  $\discIzero(t)$ and $\discIneg(t)$ at any given time $t$. Keep in mind, that states may only appear in $\discIneg(t)$ if faulty initial conditions are present or disturbances cause a state to become negative. Otherwise, it is impossible for a state to become negative.

We use the presented notation to create vectors from a selection of entries from another vector and to create matrices, which are a selection of a set of rows or columns from another matrix. For example, a vector containing all states whose indices' are given in the set $\discIpos$ is denoted as $\vecselect{\vec{\discstate}(t)}{\discIpos}$. Similarly, a matrix containing only the columns specified by the indices in the set $\discIpos$ is written as $\matcselect{\mat{A}}{\discIpos}$.

\subsection{Equilibrium Point}
\label{sec:equi_point}
First, we prove that the equilibrium point of the dynamic system depicted in \ref{fig:disc_sys} is a solution of the NNLS optimization problem \ref{eq:nnls}. To that end, we first review the KKT conditions of the NNLS problem. Let $\lsobj(\vec{\lsvar})$ refer to its objective function and $\lsineq_i(\vec{\lsvar}) = -\lsvar_i \leq 0$ for $i=1,\ldots,\discoutdim$ to its inequality constraints in standard form. Their gradients are $\nabla \lsobj(\vec{\lsvar}) = \transp{\mat{\lsmat}}\mat{\lsmat}\vec{\lsvar} - \transp{\mat{\lsmat}}\vec{\lsin}$ and $\nabla \lsineq_i(\vec{\lsvar}) = -\basisvec_{i}$ for $i=1,\ldots,\discoutdim$, respectively. Here, $\basisvec_{i}$ is the $i$-th standard basis vector. From this, we can derive the KKT conditions of problem \ref{eq:nnls} as:
\begin{equation}
\begin{aligned}
\opt{\vec{\lsvar}} &\geq \zerovec \\
\opt{\vec{\lslagmul}} &\geq \zerovec \\
\opt{\vecidx{\lslagmul}{i}} \opt{\vecidx{\lsvar}{i}} &= 0,\quad i=1,\ldots,\discoutdim \\
\transp{\mat{\lsmat}} \mat{\lsmat} \opt{\vec{\lsvar}} - \transp{\mat{\lsmat}}\vec{\lsin} - \opt{\vec{\lslagmul}} &= \zerovec.
\end{aligned}
\label{eq:nnls_kkt}
\end{equation}
Obviously, there exists a strictly feasible point, i.e., $\vec{\lsvar} > \zerovec$. Thus, Slater's condition holds for problem \ref{eq:nnls}. As a consequence, strong duality holds and the KKT conditions presented in \ref{eq:nnls_kkt} are necessary and sufficient conditions for optimality \cite{boyd_convex_2004}. Therefore, the points $\opt{\vec{\lsvar}}$ and $\opt{\vec{\lslagmul}}$ are primal and dual optimal. If the matrix $\mat{\lsmat}$ has full column-rank, the solution of problem \ref{eq:nnls} is unique \cite{bjorck_constrained_1996}.

For an equilibrium point $\vec{\discstate}_\discequi$ of a dynamic system with static input $\vec{\discin}(t) = \vec{\discin}_\discequi$
\begin{equation}
\ndot{\vec{\discstate}}(t) = \vec{\discfun}(\vec{\discstate}_\discequi,\vec{\discin}_\discequi) = \vec{0}.
\label{eq:equilibrium}
\end{equation}
must hold \cite{adamy_grundlagen_2009}.
In other words, the system states stay the same for constant input $\vec{\discin_\discequi}$, since the time derivatives of the system states vanish at the equilibrium point $\vec{\discstate}_\discequi$.

Let the input of the system described by \ref{eq:disc_sys} be constant $\vec{\discin}(t) = \vec{\lsin}$ and define
\begin{equation}
\vec{\disclagmul}_\discequi \coloneqq \transp{\mat{\discmat}}\mat{\discmat}\vec{\discstate}_\discequi - \transp{\mat{\discmat}} \vec{\lsin}.
\label{eq:disc_lag_definition}
\end{equation}
As stated by \ref{eq:equilibrium}, for an equilibrium point $\vec{\discstate}_\discequi$ it must hold for $i=1,\ldots,\discoutdim$:
\begin{equation}
0 = \vecidx{\ndot{\discstate}}{i}(t) = \vecidx{\discfun}{i}(\vec{\discstate}_\discequi,\vec{\lsin}) =
\begin{dcases}
\vecidxB{-\vec{\disclagmul}_\discequi}{i}, & \text{if } \vecidxB{\vec{\discstate}_\discequi}{i} > 0 \\
\vecidxBppart{-\vec{\disclagmul}_\discequi}{i}\!\!, & \text{if } \vecidxB{\vec{\discstate}_\discequi}{i} = 0 \\
\multicolumn{1}{c}{$\discintc,$} & \text{if } \vecidxB{\vec{\discstate}_\discequi}{i} < 0 
\end{dcases}.
\label{eq:disc_sys_equi}
\end{equation}
It can be easily seen that no $\vecidxB{\vec{\discstate}_\discequi}{i}$ can be negative since $\discintc > 0$. Thus, $\vec{\discstate}_\discequi \geq \zerovec$ must hold. For the first case of \ref{eq:disc_sys_equi}, i.e., $\vecidxB{\vec{\discstate}_\discequi}{i} > 0$, it follows that $\vecidxB{\vec{\disclagmul}_\discequi}{i} = 0$. Similarly, for the second case of \ref{eq:disc_sys_equi}, i.e., $\vecidxB{\vec{\discstate}_\discequi}{i} = 0$, we must have $\vecidxB{\vec{\disclagmul}_\discequi}{i} \geq 0$. In summary, we can write these conditions as:
\begin{equation}
\begin{aligned}
\vec{\discstate}_\discequi &\geq \zerovec \\
\vec{\disclagmul}_\discequi &\geq \zerovec \\
\vecidxB{\vec{\disclagmul}_\discequi}{i} \vecidxB{\vec{\discstate}_\discequi}{i} &= 0,\quad i=1,\ldots,\discoutdim \\
\transp{\mat{\discmat}} \mat{\discmat} \vec{\discstate}_\discequi - \transp{\mat{\discmat}}\vec{\lsin} - \vec{\disclagmul}_\discequi &= \zerovec.
\end{aligned}
\label{eq:disc_sys_kkt}
\end{equation}
Comparing \ref{eq:nnls_kkt} and \ref{eq:disc_sys_kkt}, we see that ($\vec{\discstate}_\discequi$, $\vec{\disclagmul}_\discequi$) satisfy the KKT conditions of the NNLS optimization problem and therefore are a primal and dual optimal solution. This means that given the constant system input $\vec{\discin}(t) = \vec{\lsin}$, the discontinuous system depicted in \ref{fig:disc_sys} has equilibrium points that are in fact the solution of the NNLS problem as defined in \ref{eq:nnls}.

\subsection{Caratheodory Solution}
\label{sec:caratheo_solution}
First, we reformulate system \ref{eq:disc_sys} into the following equivalent formulation:
\begin{equation}
\vecidx{\ndot{\discstate}}{i}(t) =
\left\{ \begin{array}{@{}c@{\ }l@{\,}l}
\vecidx{\discintin}{i}(t), &
\left\{ 
\vphantom{\begin{array}{@{}l}
\text{if } \vecidx{\discstate}{i}(t) > 0 \\
\text{if } \vecidx{\discstate}{i}(t) = 0,\ \vecidx{\discintin}{i}(t) \geq 0 \\
\end{array}}
\right.
&
\begin{array}{@{}l}
\text{if } \vecidx{\discstate}{i}(t) > 0 \\
\text{if } \vecidx{\discstate}{i}(t) = 0,\ \vecidx{\discintin}{i}(t) \geq 0 \\
\end{array}
\\[0.9em]
0, && \text{if } \vecidx{\discstate}{i}(t)=0,\ \vecidx{\discintin}{i}(t) < 0 \\
\discintc, && \text{if } \vecidx{\discstate}{i}(t) < 0.
\end{array}\right.
\label{eq:disc_sys_re}
\end{equation}
Here, the positive part present in the second case in \ref{eq:disc_sys} was manually split in \ref{eq:disc_sys_re}. Note also that the cases of \ref{eq:disc_sys_re} correspond to the definitions of the index sets in \ref{eq:disc_idx_sets}.

Let us investigate a time span, in which the index sets from \ref{eq:disc_idx_sets} are constant, i.e., time instances $\tau$ in an interval $\tau \in [t_j, t_{j+1})$ with $\discIpos(\tau) = \discIpos_j, \discIzero(\tau) = \discIzero_j$ and $\discIneg(\tau) = \discIneg_j$. For such a time span, the system \ref{eq:disc_sys_re} can be split into several linear subsystems. States in $\discIzero_j$ or $\discIneg_j$ are not influenced by other states, hence forming independent linear subsystems. A new linear subsystem can be formed for the states in $\discIpos_j$, which is independent of the states in $\discIzero_j$ and the influence from the states in $\discIneg_j$ can be counted as additional input, since the states in $\discIneg_j$ are independent from others. Defining
\begin{equation}
\begin{aligned}
\vec{\discstate}_j^+(\tau) &= \vecidxB{\vec{\discstate}(\tau)}{\discIpos_j}, \\
\mat{\discmat}_j^+ &= \matidxB{\mat{\discmat}}{:}{\discIpos_j}, \\
\vec{\discin}_j^+(\tau) &= \vec{\discin}(\tau) - \matidxB{\mat{\discmat}}{:}{\discIneg_j} \vecidxB{\vec{\discstate}(\tau)}{\discIneg_j},
\end{aligned}
\label{eq:disc_sys_sub}
\end{equation}
the system differential equation for this subsystem can be given as
\begin{equation}
\label{eq:disc_sys_pos}
\ndot{\vec{\discstate}}_j^+(\tau) = \mat{\discmat}_j^+ \vec{\discin}_j^+(\tau) - \transp{{\mat{\discmat}_j^+}}\!\mat{\discmat}_j^+ \vec{\discstate}_j^+(\tau).
\end{equation}
Thus, provided the matrix $\mat{\discmat}_j^+$ has full column-rank, this subsystem is globally asymptotically stable. This follows since the subsystem is linear and its system matrix $-\transp{{\mat{\discmat}_j^+}}\!\mat{\discmat}_j^+$ is negative definite, i.e., all of its eigenvalues are negative \cite{lunze_beschreibung_2013}.

Next, we inspect the transitions of states between the index sets \ref{eq:disc_idx_sets}. The index $j$, already used in the preceding paragraph, identifies these transitions and can be used to specify time intervals $[t_j, t_{j+1})$ with constant index sets. Assume $\tau \in [t_j, t_{j+1})$ in the following:
\begin{itemize}
    \item $\discIpos_j \overset{i}{\to} \discIzero_{j+1}$: When a state $i \in \discIpos_j$ in the subsystem \ref{eq:disc_sys_pos} reaches zero, it transits from $\discIpos_j$ to $\discIzero_{j+1}$. When reaching zero, the states' influence vanishes. Consequently, this transition does not cause discontinuities in any other states' trajectories.
    \item $\discIzero_j \overset{i}{\to} \discIpos_{j+1}$: If the input $\vec{\discin}(\tau)$ or changes in the values of states in $\discIpos_j$ or $\discIneg_j$ cause the term $\vecidx{\discintin}{i}(\tau)$ to become greater or equal to zero, the state $i$ transits from $\discIzero_j$ to $\discIpos_{j+1}$. However, since its immediate influence on the other states is zero, this transition is continuous.
    \item $\discIneg_j \overset{i}{\to} \discIzero_{j+1}$ or $\discIneg_j \overset{i}{\to} \discIpos_{j+1}$: At time $t_j$, the state $\vecidx{\discstate}{i}(t_j)$ must be negative, thus at constant rate of integration it will become zero at $t_{j+1} = t_j - \frac{\vecidx{\discstate}{i}(t_j)}{\discintc}$. Depending on $\vecidx{\discintin}{i}(t_{j+1})$ it will transit to either $\discIzero_{j+1}$ or $\discIpos_{j+1}$. Note, that this transition is also continuous, since the immediate influence of the state $i$ is zero.
    \item $\discIpos_j \overset{i}{\to} \discIneg_{j+1}$ or $\discIzero_j \overset{i}{\to} \discIneg_{j+1}$: These transitions are impossible with the system definition \ref{eq:disc_sys_re}.
\end{itemize}
In summary, all transitions are continuous. Provided that the system matrix $\mat{\discmat}$ has full column-rank, the behavior of the subsystem \ref{eq:disc_sys_pos} is unique at every point in time. Also, the behavior of the subsystems formed by $\discIzero_j$ and $\discIneg_j$ is unique. Finally, the transitions are obviously unique as well and thus it follows the solution $\vec{\discstate}(t)$ must be unique. Let $\discttset(t)$ be the set containing all transition times up until time $t$, i.e., $\discttset(t) = \{t_j~|~t_j \leq t\}$. Note, that $\discttset(t)$ is a countable set with measure zero. Defining $t_0 := 0$ and $t_{\card{\discttset(t)} + 1} := t$, the unique Caratheodory solution of the system follows as:
\begin{equation}
\label{eq:disc_sys_caratheo}
\vec{\discstate}(t) = \vec{\discstate}_0 + \sum_{j = 0}^{\card{\discttset(t)}} \int_{t_j}^{t_{j+1}} \!\!\!\! \vec{\discfun}(\vec{\discstate}(t'), \vec{\discin}(t')) \ud t'.
\end{equation}
The individual integrals in \ref{eq:disc_sys_caratheo} can be given using the notation of the subsystems in \ref{eq:disc_sys_sub} for $j = 1,\dots,\card{\discttset(t)}$ as:
\begin{align}
\label{eq:disc_sys_caratheo_ints}
\vecidxB{\int_{t_j}^{t_{j+1}} \!\!\!\! \dots \ud t'}{\discIpos_j} & = \int_{t_j}^{t_{j+1}} \!\!\!\! e^{-\transp{{\mat{\discmat}_{j}^+}}\!\mat{\discmat}_{j}^+ (t_{j+1} - t')} \transp{{\mat{\discmat}_{j}^+}} \vec{\discin}_j^+(t') \ud t', \nonumber \\
\vecidxB{\int_{t_j}^{t_{j+1}} \!\!\!\! \dots \ud t'}{\discIzero_j}\, & = \zerovec,\\
\vecidxB{\int_{t_j}^{t_{j+1}} \!\!\!\! \dots \ud t'}{\discIneg_j} & = \discintc \, (t_{j+1} - t_j) \, \onevec. \nonumber 
\end{align}
Thus, the Caratheodory solution is sufficiently defined. Finding a closed form of \ref{eq:disc_sys_caratheo}, however, requires expressions for the transition times and a closed form of the first equation in \ref{eq:disc_sys_caratheo_ints}, which can be very challenging.

\subsection{Stability}
\label{sec:stability}
In order to show asymptotic stability of the equilibrium point $\vec{\discstate}_\discequi$ for the system in \ref{eq:disc_sys}, we apply a linear transformation as discussed in \ref{sec:disc_dyn_sys}. We substitute $\vec{\tdiscstate}(t) = \vec{\discstate}(t) - \vec{\discstate}_\discequi$ and $\vec{\tdiscin}(t) = \vec{\discin}(t) - \vec{\discconstin}$, so that the equilibrium point of the transformed system is $\vec{\tdiscstate}_\discequi = \zerovec$ for the constant input $\vec{\tdiscin}(t) = \zerovec$. The transformed system function can be found using:
\begin{equation*}
\vecidx{\ndot{\tdiscstate}}{i}(t) = \vecidx{\tdiscfun}{i}(\vec{\tdiscstate}(t), \vec{\tdiscin}(t)) 
= \vecidx{\discfun}{i}(\vec{\tdiscstate}(t) + \vec{\discstate}_\discequi, \vec{\tdiscin}(t) + \vec{\discconstin}).
\end{equation*}
Note that we omit the time index $t$ in the following derivations to increase readability. Using \ref{eq:disc_lag_definition}, splitting the cases and simplifying, we arrive at:
\begin{align}
\label{eq:tdisc_sys}
\vecidx{\ndot{\tdiscstate}}{i}
&= \begin{dcases}
\vecidxB{-\transp{\mat{\discmat}}\mat{\discmat}\vec{\tdiscstate} -\transp{\mat{\discmat}}\mat{\discmat} \vec{\discstate}_\discequi +  \transp{\mat{\discmat}} \vec{\discconstin}}{i}, & \text{if } \vecidx{\tdiscstate}{i}+\vecidxB{\vec{\discstate}_\discequi}{i}> 0 \\
\vecidxBppart{-\transp{\mat{\discmat}}\mat{\discmat} \vec{\tdiscstate} -\transp{\mat{\discmat}}\mat{\discmat} \vec{\discstate}_\discequi +  \transp{\mat{\discmat}} \vec{\discconstin}}{i}\!\!, & \text{if } \vecidx{\tdiscstate}{i}+\vecidxB{\vec{\discstate}_\discequi}{i}=0 \\
\multicolumn{1}{c}{$\discintc,$} & \text{if } \vecidx{\tdiscstate}{i}+\vecidxB{\vec{\discstate}_\discequi}{i}<0
\end{dcases} \nonumber \\
&= \begin{dcases}
\vecidxB{-\transp{\mat{\discmat}}\mat{\discmat}\vec{\tdiscstate}}{i}, & \text{if } \vecidx{\tdiscstate}{i} + \vecidxB{\vec{\discstate}_\discequi}{i} > 0,\ \vecidxB{\vec{\discstate}_\discequi}{i} > 0 \\
\vecidxB{-\transp{\mat{\discmat}}\mat{\discmat}\vec{\tdiscstate} -\vec{\disclagmul}_\discequi}{i}, & \text{if } \vecidx{\tdiscstate}{i} > 0,\ \vecidxB{\vec{\discstate}_\discequi}{i} = 0 \\
\vecidxBppart{-\transp{\mat{\discmat}}\mat{\discmat}\vec{\tdiscstate}}{i}\!\!, & \text{if } \vecidx{\tdiscstate}{i}+\vecidxB{\vec{\discstate}_\discequi}{i}= 0, \vecidxB{\vec{\discstate}_\discequi}{i} > 0 \\
\vecidxBppart{-\transp{\mat{\discmat}}\mat{\discmat}\vec{\tdiscstate} -\vec{\disclagmul}_\discequi}{i}\!\!, & \text{if } \vecidx{\tdiscstate}{i} = 0,\ \vecidxB{\vec{\discstate}_\discequi}{i} = 0 \\
\multicolumn{1}{c}{$\discintc,$} & \text{if } \vecidx{\tdiscstate}{i}+\vecidxB{\vec{\discstate}_\discequi}{i} < 0
\end{dcases}.
\end{align}

In the following, we study stability using Lyapunov's direct method. For discontinuous dynamical systems, there exists a lot of dedicated work on stability theory, e.g., for systems with Caratheodory solutions. However, if a smooth Lyapunov function is applicable, the theory reduces to classical stability analysis \cite{cortes_discontinuous_2008}. Indeed, a smooth Lyapunov function suffices for this system and it is chosen as:
\begin{equation*}
\lyap(\vec{\tdiscstate}) = \frac{1}{2} \transp{\vec{\tdiscstate}}\vec{\tdiscstate},
\end{equation*}
with gradient $\nabla\lyap(\vec{\tdiscstate}) = \vec{\tdiscstate}$. The time derivative of $\lyap(\vec{\tdiscstate})$ yields
\begin{equation*}
\ndot{\lyap}(\vec{\tdiscstate}) = \transp{[\nabla\lyap(\vec{\tdiscstate})]}\ndot{\vec{\tdiscstate}}
= \transp{\vec{\tdiscstate}}\ndot{\vec{\tdiscstate}}
= \sum_{i = 1}^{\discstatedim} \vecidx{\tdiscstate}{i}\vecidx{\ndot{\tdiscstate}}{i}.
\end{equation*}
Three conditions must be fulfilled such that the chosen function is a Lyapunov function for the system and asymptotic stability holds:
\begin{equation}
\label{eq:lyapconditions}
    \begin{aligned}
        &\lyap(\zerovec) = 0 \,, \\
        &\lyap(\vec{\tdiscstate}) = \frac{1}{2}\norm{\vec{\tdiscstate}}{2}^2 > 0 \quad \forall \vec{\tdiscstate} \in \R^\discstatedim\setminus\{\zerovec\} \,, \\
        &\ndot{\lyap}(\vec{\tdiscstate}) < 0 \quad \forall \vec{\tdiscstate} \in \R^\discstatedim\setminus\{\zerovec\} \,.
    \end{aligned}
\end{equation}
Obviously, the first two conditions are fulfilled. With \ref{eq:tdisc_sys}, we can write
\begin{subnumcases}{\label{eq:tdisc_lyap}\vecidx{\tdiscstate}{i}\vecidx{\ndot{\tdiscstate}}{i}=}
\vecidx{\tdiscstate}{i}\vecidxB{-\transp{\mat{\discmat}}\mat{\discmat}\vec{\tdiscstate}}{i}, \hspace{-0.5cm}& if $\left\{\begin{array}{l} \vecidx{\tdiscstate}{i} + \vecidxB{\vec{\discstate}_\discequi}{i} > 0,\\ \vecidxB{\vec{\discstate}_\discequi}{i} > 0\end{array} \right.$ \label{eq:tdisc_lyapa} \\
\vecidx{\tdiscstate}{i}\vecidxB{-\transp{\mat{\discmat}}\mat{\discmat}\vec{\tdiscstate} -\vec{\disclagmul}_\discequi}{i}, \hspace{-0.5cm}& if $\left\{\begin{array}{l} \vecidx{\tdiscstate}{i}> 0,\\ \vecidxB{\vec{\discstate}_\discequi}{i} = 0\end{array} \right.$ \label{eq:tdisc_lyapb} \\
\vecidx{\tdiscstate}{i}\vecidxBppart{-\transp{\mat{\discmat}}\mat{\discmat}\vec{\tdiscstate}}{i}\!\!, \hspace{-0.5cm}& if $\left\{\begin{array}{l} \vecidx{\tdiscstate}{i}+\vecidxB{\vec{\discstate}_\discequi}{i} = 0,\\ \vecidxB{\vec{\discstate}_\discequi}{i} > 0\end{array} \right.$ \label{eq:tdisc_lyapc} \\
\vecidx{\tdiscstate}{i}\vecidxBppart{-\transp{\mat{\discmat}}\mat{\discmat}\vec{\tdiscstate} -\vec{\disclagmul}_\discequi}{i}\!\!, \hspace{-0.5cm}& if $\left\{\begin{array}{l} \vecidx{\tdiscstate}{i} = 0,\\ \vecidxB{\vec{\discstate}_\discequi}{i} = 0\end{array} \right.$ \label{eq:tdisc_lyapd} \\
\vecidx{\tdiscstate}{i}\, \discintc, \hspace{-0.5cm}& if \;\,$\vecidx{\tdiscstate}{i}+\vecidxB{\vec{\discstate}_\discequi}{i} < 0.$ \label{eq:tdisc_lyape} 
\end{subnumcases}
Note that it was shown in \ref{sec:equi_point} that $\vec{\disclagmul}_\discequi \geq 0$. For $\vec{\tdiscstate} \in \neighbourhood_\varepsilon(\zerovec)$, it follows for the subcases:
\begin{align*}
\text{(\origref{eq:tdisc_lyapb})} &= -\vecidx{\tdiscstate}{i}\vecidxB{\transp{\mat{\discmat}}\mat{\discmat}\vec{\tdiscstate}}{i} - \underbrace{\vecidx{\tdiscstate}{i}}_{\geq 0}\underbrace{\vecidxB{\vec{\disclagmul}_\discequi}{i}}_{\geq 0} \leq -\vecidx{\tdiscstate}{i}\vecidxB{\transp{\mat{\discmat}}\mat{\discmat}\vec{\tdiscstate}}{i} \\
\text{(\origref{eq:tdisc_lyapc})} &= \underbrace{\vecidx{\tdiscstate}{i}}_{= -\vecidxB{\vec{\discstate}_\discequi}{i} \leq 0} \; \underbrace{\vecidxBppart{\transp{\mat{\discmat}}\mat{\discmat}\vec{\tdiscstate}}{i}}_{\geq \vecidxB{\transp{\mat{\discmat}}\mat{\discmat}\vec{\tdiscstate}}{i}} \leq -\vecidx{\tdiscstate}{i}\vecidxB{\transp{\mat{\discmat}}\mat{\discmat}\vec{\tdiscstate}}{i} \\
\text{(\origref{eq:tdisc_lyapd})} &= \underbrace{\vecidx{\tdiscstate}{i}}_{= 0} \vecidxBppart{-\transp{\mat{\discmat}}\mat{\discmat}\vec{\tdiscstate} - \vec{\disclagmul}_\discequi}{i} = 0 =  -\underbrace{\vecidx{\tdiscstate}{i}}_{= 0} \vecidxB{\transp{\mat{\discmat}}\mat{\discmat}\vec{\tdiscstate}}{i} \\
\text{(\origref{eq:tdisc_lyape})} &= \underbrace{\vecidx{\tdiscstate}{i}}_{\leq 0} \; \underbrace{\discintc}_{\geq -\vecidxB{\transp{\mat{\discmat}}\mat{\discmat}\vec{\tdiscstate}}{i}} \leq -\vecidx{\tdiscstate}{i}\vecidxB{\transp{\mat{\discmat}}\mat{\discmat}\vec{\tdiscstate}}{i}.
\end{align*}
Thus, for $\vec{\tdiscstate} \in \R^\discstatedim\setminus\{\zerovec\}$ and $\mat{\discmat}$ having full column-rank, it holds (almost everywhere):
\begin{align}
\label{eq:lyap}
\ndot{\lyap}(\vec{\tdiscstate}) = \sum_{i = 1}^{\discoutdim} \vecidx{\tdiscstate}{i} \vecidx{\ndot{\tdiscstate}}{i} &\leq \sum_{i = 1}^{\discoutdim} -\vecidx{\tdiscstate}{i} \vecidxB{\transp{\mat{\discmat}}\mat{\discmat}\vec{\tdiscstate}}{i} \notag \\ 
&= - \transp{\vec{\tdiscstate}} \transp{\mat{\discmat}}\mat{\discmat} \vec{\tdiscstate} < 0.
\end{align}
In the last inequality, we used the fact that $\transp{\mat{\discmat}}\mat{\discmat}$ is a positive definite matrix if $\mat{\discmat}$ has full column-rank. The function $\lyap(\vec{\tdiscstate})$ satisfies the conditions \ref{eq:lyapconditions} and hence the equilibrium point $\tdiscstate = \zerovec$ is asymptotically stable in $\neighbourhood_\varepsilon(\zerovec)$. The parameter $\varepsilon$ present in the neighborhood $\neighbourhood_\varepsilon(\zerovec)$ can be chosen arbitrarily, therefore semi-global stability holds for the discontinuous system \ref{eq:disc_sys}.

\subsection{Finite Number of Switches}
\label{sec:finite_switches}
In \ref{sec:caratheo_solution}, we already established that the set of transition times $\discttset(t)$ is countable. Hence, if there is a time after which no more switches (transitions) occur, then the number of switches is finite. We show that under a mild condition on the solution, this is indeed the case. For the following analysis, we define the time-dependent vector
\begin{equation*}
\vec{\disclagmul}(t) = -\vec{\discintin}(t) = \transp{\mat{\discmat}}\mat{\discmat}\vec{\discstate}(t) - \transp{\mat{\discmat}}\vec{\discconstin},
\end{equation*}
which is the negative integrator input. Above, it was already shown that the equilibrium point $\vec{\discstate}_\discequi$ is unique if $\mat{\discmat}$ has full column-rank. Similar to \ref{eq:disc_idx_sets}, we can define unique index sets for the equilibrium point as:
\begin{equation*}
\begin{aligned}
& \discIpos_\discequi \!\!\!\!\! & &= \; \Big\{i\in\discI \,\big|\, \vecidxB{\vec{\discstate}_\discequi}{i} > 0 \Big\} \\
& & & \quad \;\; \cup \Big\{i\in\discI \,\big|\, \vecidxB{\vec{\discstate}_\discequi}{i} = 0,\ -\vecidxB{\vec{\disclagmul}_\discequi}{i} \geq 0 \Big\}, \\
& \discIzero_\discequi \!\!\!\!\! & &= \; \Big\{i\in\discI \,\big|\, \vecidxB{\vec{\discstate}_\discequi}{i} = 0,\ -\vecidxB{\vec{\disclagmul}_\discequi}{i} < 0 \Big\}, \\
& \discIneg_\discequi \!\!\!\!\! & &= \; \emptyset.
\end{aligned}
\end{equation*}
To prevent infinite oscillations at the transition boundaries, we assume there exists $\varepsilon_\discstate, \varepsilon_\disclagmul > 0$, such that
\begin{equation*}
\begin{aligned}
\forall i \in \discIpos_\discequi : \quad \vecidxB{\vec{\discstate}_\discequi}{i} &> \varepsilon_\discstate \\
\!\forall i \in \discIzero_\discequi \,: \quad \vecidxB{\vec{\disclagmul}_\discequi}{i} &> 2\varepsilon_\disclagmul.
\end{aligned}
\end{equation*}
Essentially, this means the equilibrium point $\vec{\discstate}_\discequi$ is good mannered in the sense that it is not directly on a transition boundary (see also \cite{balavoine_convergence_2012}). Note that under this assumption also 
\begin{equation*}
\begin{aligned}
\forall i \in \discIpos_\discequi : \quad \vecidxB{\vec{\disclagmul}_\discequi}{i} &= 0 \\
\!\forall i \in \discIzero_\discequi \,: \quad \vecidxB{\vec{\discstate}_\discequi}{i} &= 0
\end{aligned}
\end{equation*}
holds.

The system states $\vec{\discstate}(t)$ converge to the equilibrium point $\vec{\discstate}_\discequi$ and likewise $\vec{\disclagmul}(t)$ converges to $\vec{\disclagmul}_\discequi$. Additionally, due to the behavior of the bounded integrator \ref{eq:nnint}, after a limited time all states must be non-negative. Thus, there exists a time $\discconvtime > 0$ so that $\forall t \geq \discconvtime$:
\begin{align*}
\norm{\vec{\discstate}(t) - \vec{\discstate}_\discequi}{2} &< \varepsilon_\discstate \\
\norm{\vec{\disclagmul}(t) - \vec{\disclagmul}_\discequi}{2} &< \varepsilon_\disclagmul \\
\vec{\discstate}(t) &\geq \zerovec.
\end{align*}
Similar to the approach in \cite{balavoine_convergence_2012}, we can say that for $\forall t \geq \discconvtime$ and $\forall i \in \discIpos_\discequi$:
\begin{equation*}
\begin{gathered}
\varepsilon_\discstate > \abs{\vecidx{\discstate}{i} - \vecidxB{\vec{\discstate}_\discequi}{i}} \geq \abs{\vecidxB{\vec{\discstate}_\discequi}{i}} - \abs{\vecidx{\discstate}{i}} > \varepsilon_\discstate - \abs{\vecidx{\discstate}{i}(t)} \\
\Rightarrow \abs{\vecidx{\discstate}{i}(t)} > 0 \\
\Rightarrow \vecidx{\discstate}{i}(t) > 0.
\end{gathered}
\end{equation*}
As can be seen, for all times $t \geq \discconvtime$ the states in $\discIpos_\discequi$ stay strictly positive and hence do not switch anymore. In other words, all states in $\discIpos_\discequi$ have a certain gap to zero, that is $\vecidxB{\vec{\discstate}_\discequi}{\discIpos_\discequi}> \varepsilon_\discstate$ and after the time $\discconvtime$ all states in $\discIpos_\discequi$ stay closer to the equilibrium value than this gap.

Likewise, for $t \geq \discconvtime$ and $\forall i \in \discIzero_\discequi$ it follows:
\begin{equation*}
\begin{gathered}
\varepsilon_\disclagmul > \abs{\vecidx{\disclagmul}{i} - \vecidxB{\vec{\disclagmul}_\discequi}{i}} \geq \abs{\vecidxB{\vec{\disclagmul}_\discequi}{i}} - \abs{\vecidx{\disclagmul}{i}} > 2\varepsilon_\disclagmul - \abs{\vecidx{\disclagmul}{i}(t)} \\
\Rightarrow \abs{\vecidx{\disclagmul}{i}(t)} > \varepsilon_\disclagmul > 0 \\
\Rightarrow \vecidx{\disclagmul}{i}(t) > \varepsilon_\disclagmul > 0.
\end{gathered}
\end{equation*}
So, for all times $t \geq \discconvtime$, the integrator inputs $\vecidx{\discintin}{i}(t) = -\vecidx{\disclagmul}{i}(t)$ of the states $i \in \discIzero_\discequi$ stay strictly negative. At time $\discconvtime$, however, some states in $\discIzero$ may still be positive and some may be zero already. We investigate both cases separately:
\begin{itemize}
    \item zero states, i.e., $\forall i \in \discIzero_\discequi \cap \discIzero(\discconvtime)$: As explained above, the integrator input stays negative. Thus, no transitions occur after time $\discconvtime$.
    \item non-zero states, i.e., $\forall i \in \discIzero_\discequi \cap \discIpos(\discconvtime)$: For $t \geq \discconvtime$, it holds    
    \begin{align}
    \vecidx{\discstate}{i}(t) &= \vecidx{\discstate}{i}(\discconvtime) + \int_{\discconvtime}^{t} -\vecidx{\disclagmul}{i}(t') \ud t' \\
    &< \vecidx{\discstate}{i}(\discconvtime) + \int_{\discconvtime}^{t} -\varepsilon_\disclagmul \ud t' = \vecidx{\discstate}{i}(\discconvtime) - \varepsilon_\disclagmul (t - \discconvtime). \nonumber
    \end{align}
    Thus, a state $i$ reaches zero at time $\frac{\vecidx{\discstate}{i}(\discconvtime)}{\varepsilon_\disclagmul} + \discconvtime$ at the latest and stays zero since the integrator input remains negative. 

\end{itemize}
Hence, after time
\begin{equation*}
\max_{i \in \discIzero_\discequi \cap \discIpos(\discconvtime)} \frac{\vecidx{\discstate}{i}(\discconvtime)}{\varepsilon_\disclagmul} + \discconvtime \,,
\end{equation*}
no more switches occur. Since this point in time is finite, the number of switches is finite as well.

\subsection{Generalization to Box Constraints}
\label{sec:box_constraints}
In this section, we briefly discuss an extension to system \ref{eq:disc_sys}, so that it is able to solve optimization problems of the form
\begin{equation}
\begin{aligned}
\minimizex{\vec{\lsvar}} \frac{1}{2} \transp{\vec{\lsvar}}\mat{\lsmatB}\vec{\lsvar} - \transp{\vec{\lsinB}}\vec{\lsvar} \\
\st \vec{\lslowbox} \leq \vec{\lsvar} \leq \vec{\lshighbox}.
\end{aligned}
\label{eq:QP_lub}
\end{equation}
If the matrix $\mat{\lsmatB}$ is positive definite, which we assume in the following, this is a convex quadratic program (QP) with lower and upper bounds, see e.g. \cite{boyd_convex_2004}. By defining $\mat{\lsmatB} = \transp{\mat{\lsmat}}\mat{\lsmat}$ and $\vec{\lsinB} = \transp{\mat{\lsmat}}\vec{\lsin}$ this problem is identical to a bounded least squares problem.

The integrator used in system \ref{eq:disc_sys} is replaced by an integrator with general lower and upper bounds defined by $\vec{\lslowbox}$ and $\vec{\lshighbox}$, respectively. Also, the system matrix is replaced by the matrix $\mat{\lsmatB}$. Defining the new integrator input as $\vec{\discintinB}(t) = \vec{\discin}(t) - \mat{\lsmatB}\vec{\discstate}(t)$, we can give the system equations of the modified system as:
\begin{align}
\label{eq:disc_sys_modified}
& \vecidx{\ndot{\discstate}}{i}(t) \!\!\!\!\!\!\!\! & &= \;
\begin{dcases}
\multicolumn{1}{c}{$-\discintc,$} & \text{if } \vecidx{\discstate}{i}(t) > \vecidx{\dischighbox}{i} \\
\vecidxBnpart{\vec{\discintinB}(t)}{i}\!\!, & \text{if } \vecidx{\discstate}{i}(t) = \vecidx{\dischighbox}{i} \\
\vecidxB{\vec{\discintinB}(t)}{i}, & \text{if } \vecidx{\disclowbox}{i} < \vecidx{\discstate}{i}(t) < \vecidx{\dischighbox}{i} \\
\vecidxBppart{\vec{\discintinB}(t)}{i}\!\!, & \text{if } \vecidx{\discstate}{i}(t) = \vecidx{\disclowbox}{i} \\
\multicolumn{1}{c}{$\discintc,$} & \text{if } \vecidx{\discstate}{i}(t) < \vecidx{\disclowbox}{i}
\end{dcases} \qquad i=1,\ldots,\discoutdim \\
& \vec{\discout}(t) \!\!\!\!\!\!\!\! & &= \; \vec{\discstate}(t). \nonumber
\end{align}
If system \ref{eq:disc_sys_modified} receives the constant input $\vec{\discin}(t) = \vec{\lsinB}$, it will converge to its equilibrium point, which is the solution of the QP \ref{eq:QP_lub}. This can be shown by using the definition of an equilibrium point and comparing it to the KKT conditions of problem \ref{eq:QP_lub}. To show that a Caratheodory solution exists, one can reformulate the system equations similar to \ref{eq:disc_sys_re}. Corresponding to the cases present in the system function, five index sets (compare \ref{eq:disc_idx_sets}) can be defined: one for states which are strictly inside the bounds forming an independent linear system, two for states at the bounds and finally two for states which exceed the bounds due to initial conditions or disturbances. Transitions can be shown to be smooth and the Caratheodory solution can be characterized. Asymptotic stability holds as well. After linearly transforming the system, the same Lyapunov function as above may be used. Similarly to (\origref{eq:tdisc_lyap}), $11$ cases may be worked out and the Lyapunov function can be shown to be upper bounded by $-\transp{\vec{\tdiscstate}}\mat{\lsmatB}\vec{\tdiscstate} < 0$, as we assumed $\mat{\lsmatB}$ to be positive definite.

\section{Application: Olfactory Mixture Decomposition}
\label{sec:olf_app}
The first application of the dynamic system \ref{eq:disc_sys} we present is the decomposition of a mixture of sensory stimuli as a model for a biological neural network. It is based on the structure of the olfactory system in mammals or more precisely the main olfactory bulb (MOB). We first present the model and discuss numerical results thereafter.

\subsection{Model Description}
\label{sec:olf_model}
\begin{figure}[ht]
  \centering
  \def\svgwidth{0.75 \columnwidth}
  \footnotesize
  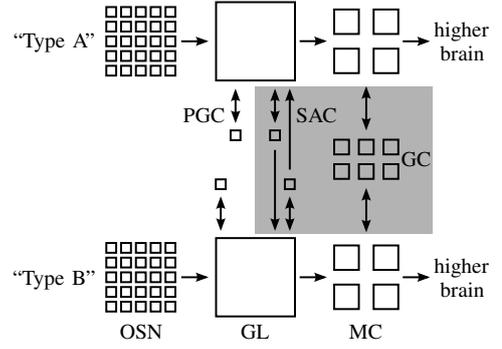
  \caption{Highly abstracted block diagram of the main olfactory bulb. Two parallel pathways, associated with two types (olfactory receptor genes) of OSNs, are shown. The sources of potential lateral interactions (SACs and GCs) are depicted within the gray shaded area. Also, the number of cells involved for each cell type is qualitatively reflected.}
  \label{fig:olf_sys}
\end{figure}

The olfactory system is considered a flat system since it consists of few processing stages before olfactory information is transmitted to higher emotional or cognitive brain regions. Olfactory sensory neurons (OSNs) are the chemical sensors of the system. By expressing different olfactory receptor genes, there are approximately $1000$ ``types'' of OSNs. There are around $1000$ to $10000$ OSNs of each type in mice~/~rodents, resulting in a total number of OSNs between $10^6$ and $10^7$, see \cite{buck_information_1996, laurent_systems_1999, mombaerts_axonal_2006, wilson_early_2006}. In the main olfactory bulb, all OSNs of the same type converge into two spherical neuropil structures of high connectivity, called glomeruli. In turn, connected to one glomerulus are approximately 25 mitral cells (MCs) \cite{buck_information_1996}, which finally project to higher brain areas. These parallel processing paths are influenced by the periglomerular cells (PGCs) and interact with each other via short axon cells (SACs) on the glomeruli level and through granule cells (GCs) on the mitral cell level, compare also \ref{fig:olf_sys}. It is believed, these so called lateral interactions play a key role in the signal processing within the MOB, see e.g. \cite{laurent_systems_1999,wilson_early_2006,murthy_olfactory_2011}. Despite the MOB's biological significance, however, the principles behind its information processing still remain unclear.

The MC $\rightarrow$ GC $\rightarrow$ MC interconnection circuit seems to be unique in the body and has a peculiar behavior. An excited MC that is connected to a GC will be inhibited via the connection (self-inhibition). Another MC, which is also connected to the same GC, will be inhibited as well. So in total the GC processing is inhibitory and thus lowers the activity of connected MCs.

Nerve cells are modeled by a dimensionless positive activity level which expresses their level of excitation. This way, we can abstract from the electro-chemical level of action potentials and do not need to make additional assumptions about the existence of rate- or timing codes, while still retaining a proportional relationship. Furthermore, the GCs are modeled as passive components and the MCs are the only active components in the system. While the GCs are able to generate action potentials, they seem to do so only rarely and may thus be covered in a passive manner by the model.

Let all the GCs and MCs in the system be identifiable by indices. Then we can model the activity level of a GC with index $j$ at a given time $t$ as follows:
\begin{equation}
    \label{eq:gc}
    \vecidx{\olfgc}{j}(t) = \sum_{i} \matidx{\olfmat}{j}{i} \vecidx{\olfmcout}{i}(t)\,,
\end{equation}
where $\vecidx{\olfmcout}{i}$ is the output of the MC with index $i$ and $\matidx{\olfmat}{j}{i} \geq 0$ is a weight describing the synaptic connection. We assume for this model that multiple GCs connected to one MC are only present to ease synaptic routing in the body. Thus in our model, through one GC a MC can be connected to all other MCs with the corresponding weights. Likewise, the $i$-th MC can be modeled as:
\begin{align}
    \label{eq:mc}
    \vecidx{\olfmcin}{i}(t) &= \vecidx{\olfglout}{i}(t) + \sum_{j} (-\matidx{\olfmat}{j}{i}) \vecidx{\olfgc}{j}(t) \nonumber \\
                            &= \vecidx{\olfglout}{i}(t) + \sum_{j} (-\matidx{\olfmat}{j}{i}) \sum_{k} \matidx{\olfmat}{j}{k} \vecidx{\olfmcout}{k}(t)\,,
\end{align}
where $\vecidx{\olfmcin}{i}$ is the input of the $i$-th MC and $\vecidx{\olfglout}{i}$ is the output of the corresponding glomerulus. Here, we assume the excitatory connections and the self-inhibitory feedback connections to have the same weights. The active MC is assumed to behave like the integrator described in \ref{eq:nnint}. Both \ref{eq:gc} and \ref{eq:mc} can be written in matrix form, using the vectors $\vec{\olfglout}$, $\vec{\olfgc}$, $\vec{\olfmcin}$ and $\vec{\olfmcout}$ which contain the output of the glomeruli, the activity levels of the GCs and the inputs and outputs of the MCs, respectively. Thus, the model of the GCs becomes $\vec{\olfgc}(t) = \mat{\olfmat} \, \vec{\olfmcout}(t)$ and the model for the MCs with interconnected GCs follows as:
\begin{equation*}
    \vec{\olfmcin}(t) = \vec{\olfglout}(t) - \transp{\mat{\olfmat}}\mat{\olfmat} \, \vec{\olfmcout}(t)\,.
\end{equation*}

One potential task of the MOB is to decompose a mixture of chemical stimuli into its individual parts. Let $\vec{\olfodour}$ be a vector of chemical stimuli of arbitrary dimension. The sensors of the olfactory system, the OSNs, react to the chemical components of this mixture and produce an output vector $\vec{\olfin}$ of dimension $\olfnumGL$. Here, we do not consider individual OSNs. Each entry of the vector can be thought of as the average output of the OSNs of one ``type'' since the purpose of the high number of individual sensors is most likely noise reduction and redundancy. Thus, the dimension $\olfnumGL$ corresponds directly to the number of glomeruli in the system.
Say the olfactory system has an internal model of odor patterns and the problem is to find the most plausible mixture that causes the current (noisy) OSN output vector $\vec{\olfin}(t)$. A smart way of solving this task would be to minimize the quadratic distance between $\vec{\olfin}(t)$ and a vector created according to the internal model. Assuming linear mixing of the chemical stimuli, the ``known'' stimuli patterns can be collected as the columns of a matrix and we assume them to be reflected in the weight matrix $\mat{\olfmat}$. If the processing of the SACs (and PGCs) on the glomerulus level can be described by $\vec{\olfglout}(t) = \transp{\mat{\olfmat}}\vec{\olfin}(t)$, then we can see that the total system behaves just like the system defined by \ref{eq:disc_sys}, explained above. Hence, under these assumptions if $\vec{\olfin}(t)$ is almost constant for a long enough time to let the dynamic system converge and the synaptic weights contain the ``known'' or ``learned'' stimuli patterns, the system solves a NNLS optimization problem to find the most likely mixture of chemical stimuli encountered. More precisely, for constant $\vec{\olfin}(t)=\vec{\olfin}$ as $t\rightarrow\infty$, $\vec{\olfmcout}(t)$ converges to the minimizer of the NNLS optimization problem \ref{eq:nnls}, with the change of variable $\mat{\lsmat} = \mat{\olfmat}$ and $\vec{\lsin} = \vec{\olfin}$. Likewise, the system equations follow by a change of variable
$\mat{\discmat} = \mat{\olfmat}$ as \ref{eq:disc_sys}.

We can make some observations stemming from the modeling. The weights in the model are non-negative, so that the weight matrix $\mat{\olfmat} \geq 0$, or equivalently $\mat{\olfmat} \in \Rnn^{\olfnumGL\times\olfnumGL}$. Assuming $\mat{\olfmat}$ is full rank, it is a basis for $\R^\olfnumGL$. It cannot, however, generate all points in $\Rnn^\olfnumGL$ using only positive linear combinations, i.e., $\vec{\dummyB} = \mat{\olfmat}\vec{\dummy}$ with $\vec{\dummy} \in \Rnn^\olfnumGL$. As said above, the vector $\vec{\olfin}$ is the noisy output of the OSNs: a noisy measurement. Say $\vec{\olfin} = \mat{\olfmat}\vec{\olfmcout}_\olftrue + \vec{\olfnoise}$, where $\vec{\olfmcout}_\olftrue$ is the actual mixture and $\vec{\olfnoise}$ is an additive noise vector. Then, although the solution $\vec{\olfnnlsopt}$ of the NNLS problem is unique and $\vec{\olfnoise}$ must be linearly dependent on $\mat{\olfmat}$, the optimal value $\frac{1}{2} \norm{\mat{\olfmat}\vec{\olfnnlsopt}-\vec{\olfin}}{2}^{2} \geq 0$. In contrast, the solution of a regular least-squares optimization problem without the non-negativity constraint would always yield an optimal value of zero in this case.

In \cite{koulakov_sparse_2011}, a similar modeling of the olfactory system is presented. However, it differs fundamentally from ours since there the GCs are modeled as active components whereas they are modeled statically here. Furthermore, a sparsity inducing cost function is assumed in \cite{koulakov_sparse_2011}, which is obsolete in our model due to the self-regularizing behavior induced by the non-negativity constraint.

\subsection{Numerical Evaluation}
\label{sec:olf_eval}
To evaluate the capabilities of the dynamic olfactory system model we performed Monte Carlo simulations. Two \emph{non-negative} data models were used in these simulations, one based on Gaussian random variables and the other based on rectangular random variables. Evidently, the performance of the dynamic olfactory system model must be evaluated for noisy data. In the following, the data models used in the Monte Carlo simulations are explained in detail. Note that we use the same data model in \ref{sec:nnsa_app} and so the data models are introduced for non-square matrices $\mat{\olfmat} \in \R^{\olfmatrowdim\times\olfnumGL}$, although $\olfmatrowdim = \olfnumGL$ in the olfactory system model.

Let $\vec{\olfmcout}_\olftrue$ be the actual mixture, having exactly $\sparsity$ nonzero entries at the indices contained in the sparse support set $\supp$. Conversely, the indices where $\vec{\olfmcout}_\olftrue$ is zero are in the complementary sparse support set, denoted as $\compsupp$. The actual (noise free) input is $\vec{\olfin}_\olftrue = \mat{\olfmat}\vec{\olfmcout}_\olftrue$. The dynamic system, however, is presented with a noisy input $\vec{\olfin} = \mat{\olfmat}\vec{\olfmcout}_\olftrue + \vec{\olfnoise}$, where $\vec{\olfnoise}$ is an additive noise vector.

For the numerical evaluation, $\vec{\olfmcout}_\olftrue$, $\mat{\olfmat}$ and $\vec{\olfnoise}$ are drawn randomly. Here, the random matrix $\mat{\olfmat}$ is entry-wise i.i.d. with expectation $\matmean$ and variance $\matvar$. The matrix $\mat{\olfmat}$ is subsequently normalized, so that the euclidean column-norm is equal to one. Thus, each entry has expectation $\expop{\matidx{\olfmat}{i}{j}} = \frac{\matmean}{\sqrt{\olfmatrowdim (\matmean^2 + \matvar)}}$ and variance $\varop{\matidx{\olfmat}{i}{j}} = \frac{\matvar}{\olfmatrowdim (\matmean^2 + \matvar)}$ after normalization. Each nonzero entry of the i.i.d. random vector $\vec{\olfmcout}_\olftrue$ has expectation $\truemixmean$ and variance $\truemixvar$. Similarly, for the i.i.d. noise vector we have $\expop{\vecidx{\olfnoise}{i}} = 0$ and $\varop{\vecidx{\olfnoise}{i}} = \noisevar$. Additionally, $\mat{\olfmat}$, $\vec{\olfmcout}_\olftrue$ and $\vec{\olfnoise}$ are entry-wise mutually independent.

We define the input signal-to-noise ratio (SNR) as $\frac{\expop{\transp{\vec{\olfin}_\olftrue}\vec{\olfin}^{}_\olftrue}}{\expop{\transp{\vec{\olfnoise}}\vec{\olfnoise}}}$. Using the definitions from above, the input SNR can be given as (a detailed derivation can be found in the Appendix):
\begin{equation}
    \label{eq:input_snr}
    \frac{\expop{\transp{\vec{\olfin}_\olftrue}\vec{\olfin}^{}_\olftrue}}{\expop{\transp{\vec{\olfnoise}}\vec{\olfnoise}}} = \frac{\sparsity}{\olfmatrowdim \noisevar} \left( \truemixvar + \frac{\truemixmean^2 \, (\sparsity \matmean^2 + \matvar)}{\matmean^2 + \matvar} \right)\,.
\end{equation}
Irrespective of the data model, the nonzero entries in $\vec{\olfmcout}_\olftrue$ are always drawn from a rectangular distribution with support $[0, \sqrt{12}]$, so that $\truemixvar = 1$ and $\truemixmean = \sqrt{3}$.

For the \emph{rectangular} (rect) data model, the entries in $\mat{\olfmat}$ and the entries in $\vec{\olfnoise}$ are drawn from rectangular distributions. This corresponds to a situation where the support of input and noise values is naturally limited by the application, for example through saturation effects. The matrix entries are drawn with support $[0, \sqrt{12}]$, so that $\matvar = 1$ and $\matmean = \sqrt{3}$. Since the noise entries are zero mean, the support interval $[-\dummy, \dummy]$ can directly be determined for the chosen input SNR by solving \ref{eq:input_snr} for $\noisevar$ and observing that $\dummy = \sqrt{3}\,\noisestd$.

In the \emph{Gaussian} model the entries are drawn from Gaussian distributions for $\mat{\olfmat}$ and $\vec{\olfnoise}$. Since the matrix $\mat{\olfmat}$ is assumed to be entry-wise non-negative, mean and variance are set to $\matmean = 5$ and $\matvar = 1$, respectively. Thus, the entries of $\mat{\olfmat}$ are non-negative with very high probability. For $\vec{\olfnoise}$, the parameter $\noisevar$ is found analogously to the rectangular case.

For the numerical evaluation, 5000 Monte Carlo instances are generated. Note that in each instance the sparse support set $\supp$, $\vec{\olfmcout}_\olftrue$, $\mat{\olfmat}$ and $\vec{\olfnoise}$ are drawn randomly as described above.

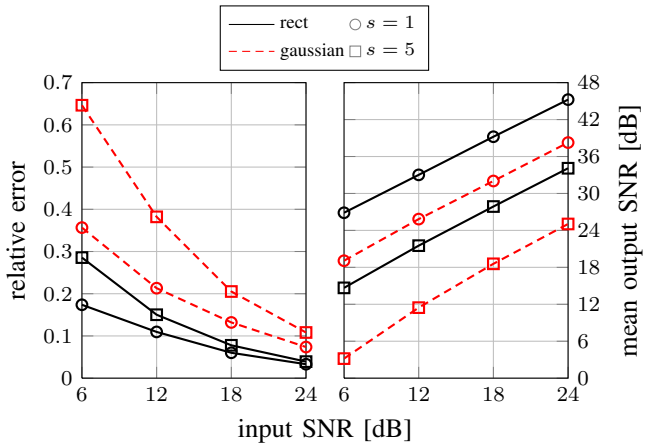
\begin{figure}[!ht]
  \centering
\makeatletter
\pgfplotsset{
	groupplot xlabel/.initial={},
	every groupplot x label/.style={
		at={($({group c1r\pgfplots@group@rows.west}|-{group c1r\pgfplots@group@rows.outer south})!0.5!({group c\pgfplots@group@columns r\pgfplots@group@rows.east}|-{group c\pgfplots@group@columns r\pgfplots@group@rows.outer south})$)},
		anchor=north,
	},
	groupplot ylabel/.initial={},
	every groupplot y label/.style={
		rotate=90,
		at={($({group c1r1.north}-|{group c1r1.outer
				west})!0.5!({group c1r\pgfplots@group@rows.south}-|{group c1r\pgfplots@group@rows.outer west})$)},
		anchor=south
	},
	execute at end groupplot/.code={%
		\node [/pgfplots/every groupplot x label]
		{\pgfkeysvalueof{/pgfplots/groupplot xlabel}};  
		\node [/pgfplots/every groupplot y label] 
		{\pgfkeysvalueof{/pgfplots/groupplot ylabel}};  
	},
	group/only outer labels/.style =
	{
		group/every plot/.code = {%
			\ifnum\pgfplots@group@current@row=\pgfplots@group@rows\else%
			\pgfkeys{xticklabels = {}, xlabel = {}}\fi%
			\ifnum\pgfplots@group@current@column=1\else%
			\pgfkeys{yticklabels = {}, ylabel = {}}\fi%
		}
	}
}

\def\endpgfplots@environment@groupplot{%
	\endpgfplots@environment@opt%
	\pgfkeys{/pgfplots/execute at end groupplot}%
	\endgroup%
}
\makeatother

\setlength\figureheight{5.5cm}
\setlength\figurewidth{4.53cm}

\begin{tikzpicture}
\begin{groupplot}[group style={group size=2 by 1, xlabels at=edge bottom, horizontal sep=0.5cm},
height=\figureheight,
width=\figurewidth,
grid=major,
xmin=6,
xmax=24,
xtick={6, 12, ..., 24},
groupplot xlabel={input SNR [dB]},
tick label style={font=\footnotesize},
legend cell align=left,
legend columns=2,
transpose legend,
legend style={at={(0.615,1.04)}, anchor=south west, font=\scriptsize}
]

\nextgroupplot[ylabel={relative error}, ymax=0.7, ymin=0, ytick={0.0, 0.1, 0.2, ..., 0.7}]

\addlegendimage{black, solid}
\addlegendentry{rect}
\addlegendimage{red, densely dashed}
\addlegendentry{gaussian}
\addlegendimage{mark=o, only marks, gray}
\addlegendentry{$\sparsity = 1$}
\addlegendimage{mark=square, only marks, gray}
\addlegendentry{$\sparsity = 5$}


\addplot [
color=black,
thick,
solid,
mark=o,
mark options=solid
]
table {\figpath mean_rel_err_uni_olf_s1.dat};

\addplot [
color=black,
thick,
solid,
mark=square,
mark options=solid
]
table {\figpath mean_rel_err_uni_olf_s5.dat};


\addplot [
color=red,
thick,
densely dashed,
mark=o,
mark options=solid
]
table {\figpath mean_rel_err_gauss_olf_s1.dat};

\addplot [
color=red,
thick,
densely dashed,
mark=square,
mark options=solid
]
table {\figpath mean_rel_err_gauss_olf_s5.dat};

\nextgroupplot[ylabel={mean output SNR [dB]}, yticklabel pos=right, ymin=0, ymax=48, ytick={0, 6, 12, 18, ..., 48}]


\addplot [
color=black,
thick,
solid,
mark=o,
mark options=solid
]
table {\figpath mean_SNR_out_uni_olf_s1.dat};

\addplot [
color=black,
thick,
solid,
mark=square,
mark options=solid
]
table {\figpath mean_SNR_out_uni_olf_s5.dat};


\addplot [
color=red,
thick,
densely dashed,
mark=o,
mark options=solid
]
table {\figpath mean_SNR_out_gauss_olf_s1.dat};

\addplot [
color=red,
thick,
densely dashed,
mark=square,
mark options=solid
]
table {\figpath mean_SNR_out_gauss_olf_s5.dat};

\makeatletter

\end{groupplot}
\end{tikzpicture}
  \caption{\emph{left:} Mean relative error on the sparse support set $\supp$ for different input SNRs. \emph{right:} Mean output SNR (see \ref{eq:output_snr}) for different input SNRs.}
  \label{fig:olf_rel_err_output_snr}
\end{figure}

In \ref{fig:olf_rel_err_output_snr}, on the left the mean relative error on the sparse support set $\supp$ is plotted for different input SNRs as well as different sparsities $\sparsity$ for both non-negative data models. As can be seen, higher sparsities correspond to higher errors, which is to be expected. Also, the Gaussian data model generally performs worse than the rectangular one. Though relative errors of less than $10\%$ can be reached, when the input SNR is high enough.

To show the denoising capabilities of this system, we evaluate the output SNR, which is defined as:
\begin{equation}
\label{eq:output_snr}
\expop{\frac{\transp{\vecselect{\vec{\olfmcout}}{\supp}}\vecselect{\vec{\olfmcout}}{\supp}^{}}{\transp{\vecselect{\vec{\olfmcout}}{\compsupp}}\vecselect{\vec{\olfmcout}}{\compsupp}^{}}} \,.
\end{equation}
There, $\vecselect{\vec{\olfmcout}}{\supp}$ is a vector comprised only of the entries of $\vec{\olfmcout}$ whose indices are contained in the set $\supp$. Hence, the output SNR is the ratio between the power in the sparse support $\supp$ (desired signal) and the power in the complementary sparse support $\compsupp$ (noise) after recovery. The mean output SNR versus the input SNR is shown in \ref{fig:olf_rel_err_output_snr} on the right side. For both data models under low sparsities $\sparsity$, the output SNR is significantly higher than the input SNR. This suggests that the system possesses denoising capabilities.

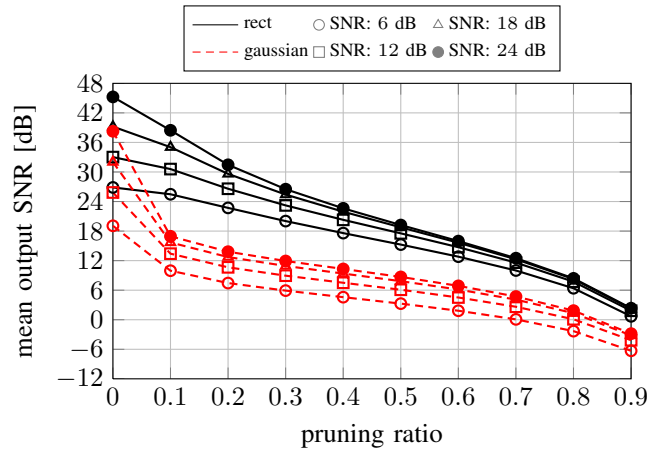
\begin{figure}[!ht]
  \centering
  \setlength\figureheight{5.5cm}
\setlength\figurewidth{8.4cm}

\begin{tikzpicture}
\begin{axis}[%
width=\figurewidth,
height=\figureheight,
grid=major,
xmin=0,
xmax=0.9,
xtick={0, 0.1, 0.2, ..., 0.9},
ymin=-12,
ymax=48,
ytick={-12, -6, 0, 6, ..., 48},
ylabel={mean output SNR [dB]},
xlabel={pruning ratio},
legend cell align=left,
legend style={at={(0.5,1.04)}, anchor=south, font=\scriptsize},
legend entries={rect, gaussian, SNR: $6~\dB$, SNR: $12~\dB$, SNR: $18~\dB$, SNR: $24~\dB$},
legend columns=2,
transpose legend
]

\addlegendimage{black, solid}
\addlegendimage{red, densely dashed}
\addlegendimage{mark=o, only marks, gray}
\addlegendimage{mark=square, only marks, gray}
\addlegendimage{mark=triangle, only marks, gray}
\addlegendimage{mark=*, only marks, gray}


\addplot [
color=black,
thick,
solid,
mark=*,
mark options=solid
]
table{\figpath mean_SNR_out_uni_olf_s_ratio_snr24_s1.dat};

\addplot [
color=black,
thick,
solid,
mark=triangle,
mark options=solid
]
table{\figpath mean_SNR_out_uni_olf_s_ratio_snr18_s1.dat};

\addplot [
color=black,
thick,
solid,
mark=square,
mark options=solid
]
table{\figpath mean_SNR_out_uni_olf_s_ratio_snr12_s1.dat};

\addplot [
color=black,
thick,
solid,
mark=o,
mark options=solid
]
table{\figpath mean_SNR_out_uni_olf_s_ratio_snr6_s1.dat};


\addplot [
color=red,
thick,
densely dashed,
mark=*,
mark options=solid
]
table{\figpath mean_SNR_out_gauss_olf_s_ratio_snr24_s1.dat};

\addplot [
color=red,
thick,
densely dashed,
mark=triangle,
mark options=solid
]
table{\figpath mean_SNR_out_gauss_olf_s_ratio_snr18_s1.dat};

\addplot [
color=red,
thick,
densely dashed,
mark=square,
mark options=solid
]
table{\figpath mean_SNR_out_gauss_olf_s_ratio_snr12_s1.dat};

\addplot [
color=red,
thick,
densely dashed,
mark=o,
mark options=solid
]
table{\figpath mean_SNR_out_gauss_olf_s_ratio_snr6_s1.dat};

\end{axis}
\end{tikzpicture}
  \caption{Mean output SNR for different pruning ratios of the matrix $\mat{\olfmat}$.}
  \label{fig:olf_output_snr_pruning}
\end{figure}

Biological systems exhibit high fault tolerance and tend to be power efficient. Since the matrix $\mat{\olfmat}$ models synaptic connections, we can test two effects by setting the smallest entries in the matrix to zero. On the one hand, we may simulate synaptic failures this way and thereby test the fault tolerance of the system. On the other hand, we may also test robustness and power efficient operation at the same time since maintaining a multitude of synaptic connections comes at a high cost for the body. In \ref{fig:olf_output_snr_pruning}, we show the mean output SNR evaluated for pruning ratios between $0$ and $0.9$. Here, a pruning ratio of $0.1$ means that the smallest $10\%$ of the matrix entries of $\mat{\olfmat}$ are set to zero. Especially for the rectangular data model, we can see that even moderate pruning ratios ($\approx 0.5 - 0.7$) are acceptable since the output SNR stays higher than the input SNR.

In summary, these results indicate that a topologically simple neural network like the olfactory system could in principle tackle complicated signal recovery tasks, e.g., non-negative sparse approximation. It could greatly benefit from the inherent denoising capability and the fault tolerance operation of the presented network.

\section{Application: Non-Negative Sparse Approximation}
\label{sec:nnsa_app}
The second application we present is solving non-negative sparse approximation problems. Here, we discuss a more general model compared to the previous section including the very important class of underdetermined problems. We first introduce the model and present numerical evaluations afterwards.

\subsection{Model description}
\label{sec:nnsa_model}
We consider non-negative sparse approximation problems as an application of the network presented in \ref{sec:system_model}. As introduced similarly in \ref{sec:cssa}, the task is to recover an $\sparsity$-sparse signal $\vec{\cssignal}_\cstrue$ from linear measurements. The non-zero entries of $\vec{\cssignal}_\cstrue$ are contained in the sparse support set $\supp$ and it is assumed that these entries are non-negative, i.e., $\vecidxB{\vec{\cssignal}_\cstrue}{\supp} \in \Rp^{\card{\supp}}$, where $\card{\supp}$ is the cardinality of the set $\supp$. The measured signal $\vec{\csmeas}$ is obtained from a noisy linear measurement process, which is described by $\vec{\csmeas} = \mat{\csmat}\vec{\cssignal}_\cstrue + \vec{\csmeaserr}$ with $\mat{\csmat} \in \Rp^{\csmeasdim\times\cssignaldim}$ and $\vec{\csmeaserr} \in \R^{\csmeasdim}$. Typically, the number of measurements is assumed to be small so that $\csmeasdim \ll \cssignaldim$. Note that we assume for simplicity that all entries of the matrix $\mat{\csmat}$ are strictly greater than zero. This implies that the matrix fulfills the self-regularizing property \cite{slawski_non-negative_2013} (see also \ref{sec:cssa}). Furthermore, we assume that the measured signal is element-wise non-negative, i.e., $\vec{\csmeas} \geq 0$, with very high probability. This holds if the noise $\vec{\csmeaserr}$ is zero mean, the distribution of $\vec{\csmeaserr}$ has finite support or tails with negligible contributions and the SNR is high enough.

To solve non-negative sparse approximation problems with the discontinuous dynamic system from \ref{sec:system_model}, the system is presented with the measurement at its input $\vec{\discin}(t) = \vec{\csmeas}$ and the system matrix equals the linear system model. The systems output $\vec{\discout}(t)$ will hold the recovered signal vector $\vec{\cssignal}$, once it has converged.

In the convergence analysis in \ref{sec:conv_analysis}, in some places it was required that the system matrix $\mat{\discmat}$ has full column-rank. Obviously, if $\csmeasdim < \cssignaldim$, as is typical in non-negative sparse approximation problems, the matrix cannot have full column-rank by definition. We discuss the consequences in the following. We see, that it does not affect the practical applicability of the network for the relevant problem class. 

The solution of the NNLS problem is not unique if the matrix $\mat{\discmat}$ does not have full column-rank (compare \ref{sec:equi_point}). However, the KKT conditions of the discontinuous dynamic system and the NNLS problem are still equal. Moreover, since we assume that the matrix $\mat{\discmat}$ is entry-wise non-negative, the row-span of the matrix intersects the positive orthant and the self-regularizing property is fulfilled \cite{slawski_non-negative_2013}. Thus, if no noise is present the solution is unique nevertheless and if noise is present, the recovery error is bounded. So even if the solution of the NNLS problem would differ from the equilibrium point of the dynamic system in the noisy case, because the KKT conditions are equal, the error is bounded for both solutions. 

For a given starting point $\vec{\discstate}(0)$ and a given constant input $\vec{\discin}(t) = \vec{\csmeas}$, we can still split the system into linear subsystems (compare \ref{sec:caratheo_solution}). The ``active'' subsystems corresponding to the index sets $\discIpos_j$ are (locally) asymptotically stable since $- \transp{\vec{\tdiscstate}} \transp{{\mat{\discmat}_j^+}}\mat{\discmat}_j^+ \vec{\tdiscstate} < 0$ for all $\vec{\tdiscstate} \neq \zerovec$, due to the fact that $\mat{\discmat}_j^+ > 0$. Hence, all state transitions remain unique and continuous. Also, the Caratheodory solution can be defined in the same way as before. The stability analysis can be performed using a given starting point $\vec{\discstate}(0)$, a given constant input $\vec{\discin}(t) = \vec{\csmeas}$ and a corresponding equilibrium point $\vec{\discstate}_\discequi$ (which depends on $\vec{\discstate}(0)$ and $\vec{\csmeas}$). Since $\mat{\discmat} > 0$, \ref{eq:lyap} still holds although $\transp{\mat{\discmat}}\mat{\discmat}$ is only positive semi-definite. For a given tuple $(\vec{\discstate}(0),\vec{\discin}(t) = \vec{\csmeas}, \vec{\discstate}_\discequi)$ and the condition on the solution the proof of the finite number of switches during convergence remains the same as well (c.f. \ref{sec:finite_switches}). 

Summarizing, even for the case that $\csmeasdim < \cssignaldim$ which prevents the matrix $\mat{\csmat}$ from having full column-rank, the system is (locally) asymptotically stable if $\mat{\csmat} > 0$ and can thus be used to solve non-negative sparse approximation problems.

\subsection{Numerical Evaluation}
\label{nnsa_eval}
For the evaluation, we use the same non-negative data models introduced in \ref{sec:olf_eval}. Here, our main focus of the evaluation is to compare the recovery performance of the dynamic system (NNLS) with non-negative basis pursuit denoising (NNBPDN):
\begin{equation}
    \label{eq:nnbpdn}
    \begin{aligned}
        \minimizex{\vec{\cssignal}} \frac{1}{2}\norm{\mat{\csmat}\vec{\cssignal} - \vec{\csmeas}}{2}^2 + \csreg \norm{\vec{\cssignal}}{1} \\
        \st \vec{\cssignal} \geq \zerovec\,.
    \end{aligned}
\end{equation}
Note that in contrast to the NNLS problem, the NNBPDN problem has a regularization parameter $\csreg$ which must be determined in advance. For the evaluation, for each Monte Carlo instance, we solve the NNBPDN problem for $50$ logarithmically spaced regularization parameters and pick the solution which has the smallest mean squared error (MSE) with respect to the actual signal vector $\vec{\cssignal}_\cstrue$. This gives the NNBPDN problem an advantage since in practice one obviously cannot use this procedure and $\csreg$ must be fixed to some empirically determined value beforehand.

\begin{figure}[!ht]
  \centering
  \setlength\figureheight{5.5cm}
\setlength\figurewidth{8.3cm}

\begin{tikzpicture}
\begin{semilogyaxis}[%
width=\figurewidth,
height=\figureheight,
grid=major,
xmin=25,
xmax=200,
xtick={25, 50, 75, ..., 200},
ymin=1e-4,
ymax=1e1,
ylabel={MSE on support},
xlabel={$\#$ measurements ($\discindim$)},
legend cell align=left,
legend style={at={(0.5,1.04)}, anchor=south, font=\scriptsize},
legend entries={NNBPDN, NNLS, SNR: $10~\dB$, SNR: $20~\dB$, SNR: $30~\dB$, SNR: $40~\dB$},
legend columns=2,
transpose legend
]

\addlegendimage{black, solid}
\addlegendimage{red, densely dashed}
\addlegendimage{mark=o, only marks, gray}
\addlegendimage{mark=square, only marks, gray}
\addlegendimage{mark=triangle, only marks, gray}
\addlegendimage{mark=*, only marks, gray}


\addplot [
color=black,
thick,
solid,
mark=o,
mark options=solid
]
table{\figpath avg_mse_sup_NNBPDN_uni_s_5_snr_10.dat};

\addplot [
color=red,
thick,
densely dashed,
mark=o,
mark options=solid
]
table{\figpath avg_mse_sup_NNLS_uni_s_5_snr_10.dat};

\addplot [
color=black,
thick,
solid,
mark=square,
mark options=solid
]
table{\figpath avg_mse_sup_NNBPDN_uni_s_5_snr_20.dat};

\addplot [
color=red,
thick,
densely dashed,
mark=square,
mark options=solid
]
table{\figpath avg_mse_sup_NNLS_uni_s_5_snr_20.dat};

\addplot [
color=black,
thick,
solid,
mark=triangle,
mark options=solid
]
table{\figpath avg_mse_sup_NNBPDN_uni_s_5_snr_30.dat};

\addplot [
color=red,
thick,
densely dashed,
mark=triangle,
mark options=solid
]
table{\figpath avg_mse_sup_NNLS_uni_s_5_snr_30.dat};

\addplot [
color=black,
thick,
solid,
mark=*,
mark options=solid
]
table{\figpath avg_mse_sup_NNBPDN_uni_s_5_snr_40.dat};

\addplot [
color=red,
thick,
densely dashed,
mark=*,
mark options=solid
]
table{\figpath avg_mse_sup_NNLS_uni_s_5_snr_40.dat};

\end{semilogyaxis}
\end{tikzpicture}
  \caption{Comparison of the MSE with respect to the actual signal vector $\vec{\cssignal}_\cstrue$ on the sparse support set $\supp$ for different input SNRs, $\cssignaldim = 200$, $\sparsity = 5$ and using the uniform (rect) data model. Note the logarithmic scaling of the ordinate.}
  \label{fig:mse_sup}
\end{figure}
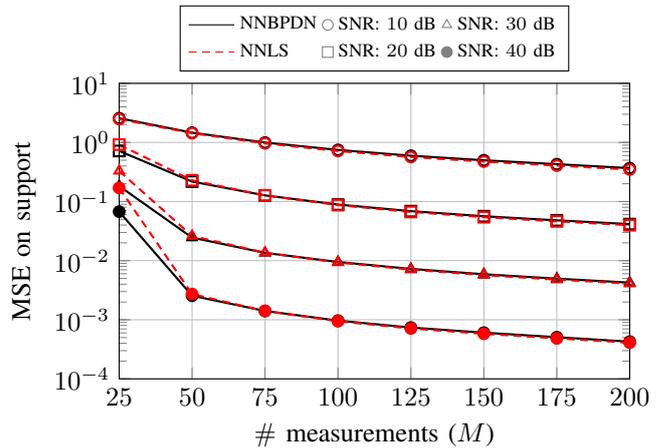

In \ref{fig:mse_sup}, the MSE of the recovered signal $\vec{\cssignal}$ with respect to the actual signal vector $\vec{\cssignal}_\cstrue$ on the indices contained in $\supp$, i.e., the MSE on the sparse support, is shown for different input SNRs. As can be seen, both methods differ only marginally with NNLS featuring a slightly lower error for a medium to high number of measurements $\csmeasdim$. Only for a low number of measurements and higher SNRs, NNBPDN has a smaller error on average. 

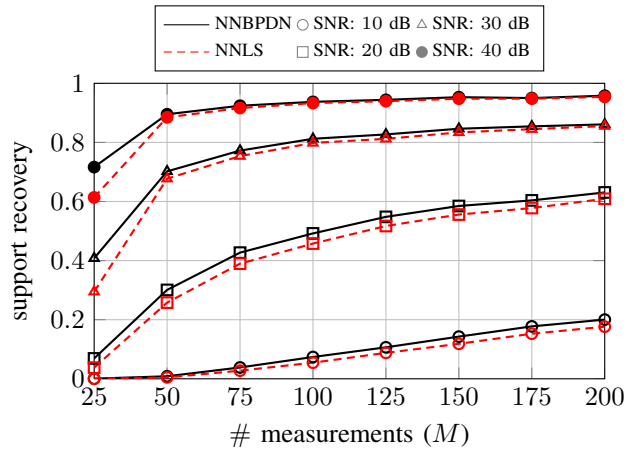
\begin{figure}[!ht]
  \centering
  \setlength\figureheight{5.5cm}
\setlength\figurewidth{8.3cm}

\begin{tikzpicture}
\begin{axis}[%
width=\figurewidth,
height=\figureheight,
grid=major,
xmin=25,
xmax=200,
xtick={25, 50, 75, ..., 200},
ymin=0,
ymax=1,
ylabel={support recovery},
xlabel={$\#$ measurements ($\discindim$)},
legend cell align=left,
legend style={
at={(0.5,1.04)},
anchor=south, font=\scriptsize},
legend entries={NNBPDN, NNLS, SNR: $10~\dB$, SNR: $20~\dB$, SNR: $30~\dB$, SNR: $40~\dB$},
legend columns=2,
transpose legend
]

\addlegendimage{black, solid}
\addlegendimage{red, densely dashed}
\addlegendimage{mark=o, only marks, gray}
\addlegendimage{mark=square, only marks, gray}
\addlegendimage{mark=triangle, only marks, gray}
\addlegendimage{mark=*, only marks, gray}


\addplot [
color=black,
thick,
solid,
mark=o,
mark options=solid
]
table{\figpath ssr_NNBPDN_uni_s_5_snr_10.dat};

\addplot [
color=red,
thick,
densely dashed,
mark=o,
mark options=solid
]
table{\figpath ssr_NNLS_uni_s_5_snr_10.dat};

\addplot [
color=black,
thick,
solid,
mark=square,
mark options=solid
]
table{\figpath ssr_NNBPDN_uni_s_5_snr_20.dat};

\addplot [
color=red,
thick,
densely dashed,
mark=square,
mark options=solid
]
table{\figpath ssr_NNLS_uni_s_5_snr_20.dat};

\addplot [
color=black,
thick,
solid,
mark=triangle,
mark options=solid
]
table{\figpath ssr_NNBPDN_uni_s_5_snr_30.dat};

\addplot [
color=red,
thick,
densely dashed,
mark=triangle,
mark options=solid
]
table{\figpath ssr_NNLS_uni_s_5_snr_30.dat};

\addplot [
color=black,
thick,
solid,
mark=*,
mark options=solid
]
table{\figpath ssr_NNBPDN_uni_s_5_snr_40.dat};

\addplot [
color=red,
thick,
densely dashed,
mark=*,
mark options=solid
]
table{\figpath ssr_NNLS_uni_s_5_snr_40.dat};

\end{axis}
\end{tikzpicture}
  \caption{Fraction of monte carlo instances in which clear recovery of the sparse support set $\supp$ is possible through thresholding for different input SNRs, $\cssignaldim = 200$, $\sparsity = 5$ and using the uniform (rect) data model.}
  \label{fig:ssr}
\end{figure}

Next, we evaluate the sparse support set recovery performance in \ref{fig:ssr}. Here, we check whether there exists a threshold with which the sparse support $\supp$ can be separated from the complementary sparse support $\compsupp$, i.e., whether $\max(\vecselect{\vec{\cssignal}}{\compsupp}) < \min(\vecselect{\vec{\cssignal}}{\supp})$. We can see, that NNBPDN seems to be slightly superior as far as recovery of the support is concerned.

\begin{figure}[!ht]
  \centering
\makeatletter
\pgfplotsset{
    groupplot xlabel/.initial={},
    every groupplot x label/.style={
        at={($({group c1r\pgfplots@group@rows.west}|-{group c1r\pgfplots@group@rows.outer south})!0.5!({group c\pgfplots@group@columns r\pgfplots@group@rows.east}|-{group c\pgfplots@group@columns r\pgfplots@group@rows.outer south})$)},
        anchor=north,
    },
    groupplot ylabel/.initial={},
    every groupplot y label/.style={
            rotate=90,
        at={($({group c1r1.north}-|{group c1r1.outer
west})!0.5!({group c1r\pgfplots@group@rows.south}-|{group c1r\pgfplots@group@rows.outer west})$)},
        anchor=south
    },
    execute at end groupplot/.code={%
      \node [/pgfplots/every groupplot x label]
{\pgfkeysvalueof{/pgfplots/groupplot xlabel}};  
      \node [/pgfplots/every groupplot y label] 
{\pgfkeysvalueof{/pgfplots/groupplot ylabel}};  
    },
    group/only outer labels/.style =
{
group/every plot/.code = {%
    \ifnum\pgfplots@group@current@row=\pgfplots@group@rows\else%
        \pgfkeys{xticklabels = {}, xlabel = {}}\fi%
    \ifnum\pgfplots@group@current@column=1\else%
        \pgfkeys{yticklabels = {}, ylabel = {}}\fi%
}
}
}

\def\endpgfplots@environment@groupplot{%
    \endpgfplots@environment@opt%
    \pgfkeys{/pgfplots/execute at end groupplot}%
    \endgroup%
}
\makeatother

\setlength\figureheight{5.5cm}
\setlength\figurewidth{4.48cm}

\begin{tikzpicture}
\begin{groupplot}[group style={group size=2 by 1, xlabels at=edge bottom, horizontal sep=0.5cm},
height=\figureheight,
width=\figurewidth,
grid=major,
xmin=5,
xmax=20,
xtick={5, 10, 15, 20},
groupplot xlabel={sparsity ($\sparsity$)},
tick label style={font=\footnotesize},
legend cell align=left,
legend columns=2,
transpose legend,
legend style={at={(0.19,1.04)}, anchor=south west, font=\scriptsize}
]

\nextgroupplot[ylabel={MSE on support}, ymax=4, ymin=0]

\addlegendimage{black, solid}
\addlegendentry{NNBPDN}
\addlegendimage{red, densely dashed}
\addlegendentry{NNLS}
\addlegendimage{mark=o, only marks, gray}
\addlegendentry{SNR: $10~\dB$}
\addlegendimage{mark=square, only marks, gray}
\addlegendentry{SNR: $20~\dB$}
\addlegendimage{mark=triangle, only marks, gray}
\addlegendentry{SNR: $30~\dB$}
\addlegendimage{mark=*, only marks, gray}
\addlegendentry{SNR: $40~\dB$}


\addplot [
color=black,
thick,
solid,
mark=o,
mark options=solid
]
table{\figpath avg_mse_sup_NNBPDN_uni_m_50_snr_10.dat};

\addplot [
color=red,
thick,
densely dashed,
mark=o,
mark options=solid
]
table{\figpath avg_mse_sup_NNLS_uni_m_50_snr_10.dat};

\addplot [
color=black,
thick,
solid,
mark=square,
mark options=solid
]
table{\figpath avg_mse_sup_NNBPDN_uni_m_50_snr_20.dat};

\addplot [
color=red,
thick,
densely dashed,
mark=square,
mark options=solid
]
table{\figpath avg_mse_sup_NNLS_uni_m_50_snr_20.dat};

\addplot [
color=black,
thick,
solid,
mark=triangle,
mark options=solid
]
table{\figpath avg_mse_sup_NNBPDN_uni_m_50_snr_30.dat};

\addplot [
color=red,
thick,
densely dashed,
mark=triangle,
mark options=solid
]
table{\figpath avg_mse_sup_NNLS_uni_m_50_snr_30.dat};

\addplot [
color=black,
thick,
solid,
mark=*,
mark options=solid
]
table{\figpath avg_mse_sup_NNBPDN_uni_m_50_snr_40.dat};

\addplot [
color=red,
thick,
densely dashed,
mark=*,
mark options=solid
]
table{\figpath avg_mse_sup_NNLS_uni_m_50_snr_40.dat};

\nextgroupplot[ylabel={support recovery}, yticklabel pos=right, ymin=0, ymax=1]


\addplot [
color=black,
thick,
solid,
mark=o,
mark options=solid
]
table{\figpath ssr_NNBPDN_uni_m_50_snr_10.dat};

\addplot [
color=red,
thick,
densely dashed,
mark=o,
mark options=solid
]
table{\figpath ssr_NNLS_uni_m_50_snr_10.dat};

\addplot [
color=black,
thick,
solid,
mark=square,
mark options=solid
]
table{\figpath ssr_NNBPDN_uni_m_50_snr_20.dat};

\addplot [
color=red,
thick,
densely dashed,
mark=square,
mark options=solid
]
table{\figpath ssr_NNLS_uni_m_50_snr_20.dat};

\addplot [
color=black,
thick,
solid,
mark=triangle,
mark options=solid
]
table{\figpath ssr_NNBPDN_uni_m_50_snr_30.dat};

\addplot [
color=red,
thick,
densely dashed,
mark=triangle,
mark options=solid
]
table{\figpath ssr_NNLS_uni_m_50_snr_30.dat};

\addplot [
color=black,
thick,
solid,
mark=*,
mark options=solid
]
table{\figpath ssr_NNBPDN_uni_m_50_snr_40.dat};

\addplot [
color=red,
thick,
densely dashed,
mark=*,
mark options=solid
]
table{\figpath ssr_NNLS_uni_m_50_snr_40.dat};

\end{groupplot}
\end{tikzpicture}
  \caption{Influence of the sparsity $\sparsity$ for different input SNRs, $\csmeasdim = 50$, $\cssignaldim = 200$ for the uniform (rect) data model. \emph{left:} MSE on sparse support $\supp$. \emph{right:} sparse support recovery by thresholding.}
  \label{fig:mse_sup_ssr_combi_over_s}
\end{figure}
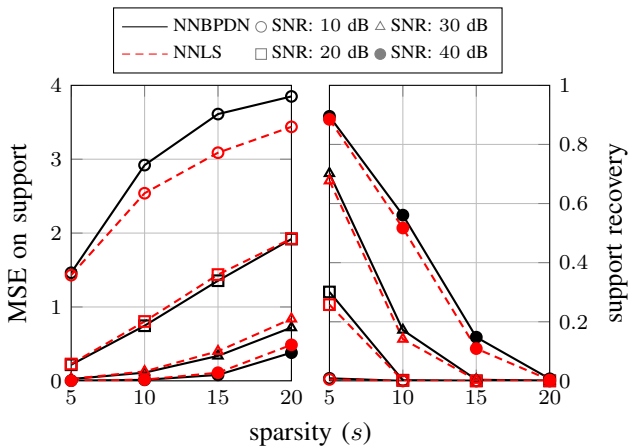

To analyze the influence of the sparsity on the recovery performance, \ref{fig:mse_sup_ssr_combi_over_s} depicts the MSE on the support on the left and the support recovery fraction on the right both for different values of the sparsity $\sparsity$. As expected, the higher the sparsity the higher the error and hence also the lower the fraction of recoverable supports. Although NNBPDN performs slightly superior for most choices of the parameters, it is interesting to note that for low SNR situations the discontinuous network yields solutions that have a lower MSE on the support on average.

\begin{figure}[!ht]
  \centering
\makeatletter
\pgfplotsset{
    groupplot xlabel/.initial={},
    every groupplot x label/.style={
        at={($({group c1r\pgfplots@group@rows.west}|-{group c1r\pgfplots@group@rows.outer south})!0.5!({group c\pgfplots@group@columns r\pgfplots@group@rows.east}|-{group c\pgfplots@group@columns r\pgfplots@group@rows.outer south})$)},
        anchor=north,
    },
    groupplot ylabel/.initial={},
    every groupplot y label/.style={
            rotate=90,
        at={($({group c1r1.north}-|{group c1r1.outer
west})!0.5!({group c1r\pgfplots@group@rows.south}-|{group c1r\pgfplots@group@rows.outer west})$)},
        anchor=south
    },
    execute at end groupplot/.code={%
      \node [/pgfplots/every groupplot x label]
{\pgfkeysvalueof{/pgfplots/groupplot xlabel}};  
      \node [/pgfplots/every groupplot y label] 
{\pgfkeysvalueof{/pgfplots/groupplot ylabel}};  
    },
    group/only outer labels/.style =
{
group/every plot/.code = {%
    \ifnum\pgfplots@group@current@row=\pgfplots@group@rows\else%
        \pgfkeys{xticklabels = {}, xlabel = {}}\fi%
    \ifnum\pgfplots@group@current@column=1\else%
        \pgfkeys{yticklabels = {}, ylabel = {}}\fi%
}
}
}

\def\endpgfplots@environment@groupplot{%
    \endpgfplots@environment@opt%
    \pgfkeys{/pgfplots/execute at end groupplot}%
    \endgroup%
}
\makeatother

\setlength\figureheight{5.5cm}
\setlength\figurewidth{4.23cm}

\begin{tikzpicture}
\begin{groupplot}[%
group style={group size=2 by 1, xlabels at=edge bottom, horizontal sep=0.5cm},
height=\figureheight,
width=\figurewidth,
grid=major,
xmin=25,
xmax=200,
xtick={25, 50, 100, ..., 200},
groupplot xlabel={$\#$ measurements ($\discindim$)},
tick label style={font=\footnotesize},
legend cell align=left,
legend columns=2,
transpose legend,
legend style={at={(0.12,1.04)}, anchor=south west, font=\scriptsize}
]

\nextgroupplot[ylabel={MSE on support}, ymode=log, ymax=1, ymin=1e-4]

\addlegendimage{black, solid}
\addlegendentry{NNBPDN rect}
\addlegendimage{red, densely dashed}
\addlegendentry{NNLS rect}
\addlegendimage{black, densely dotted}
\addlegendentry{NNBPDN gaussian}
\addlegendimage{red, densely dashdotted}
\addlegendentry{NNLS gaussian}

\addplot [
color=black,
thick,
solid,
mark=*,
mark options=solid,
mark size=1pt
]
table{\figpath avg_mse_sup_NNBPDN_uni_s_5_snr_40.dat};

\addplot [
color=red,
thick,
densely dashed,
mark=*,
mark options=solid,
mark size=1pt
]
table{\figpath avg_mse_sup_NNLS_uni_s_5_snr_40.dat};

\addplot [
color=black,
thick,
densely dotted,
mark=*,
mark options=solid,
mark size=1pt
]
table{\figpath avg_mse_sup_NNBPDN_gauss_s_5_snr_40.dat};

\addplot [
color=red,
thick,
densely dashdotted,
mark=*,
mark options=solid,
mark size=1pt
]
table{\figpath avg_mse_sup_NNLS_gauss_s_5_snr_40.dat};

\nextgroupplot[ylabel={support recovery}, yticklabel pos=right, ymin=0.2, ymax=1]

\addplot [
color=black,
thick,
solid,
mark=*,
mark options=solid,
mark size=1pt
]
table{\figpath ssr_NNBPDN_uni_s_5_snr_40.dat};

\addplot [
color=red,
thick,
densely dashed,
mark=*,
mark options=solid,
mark size=1pt
]
table{\figpath ssr_NNLS_uni_s_5_snr_40.dat};

\addplot [
color=black,
thick,
densely dotted,
mark=*,
mark options=solid,
mark size=1pt
]
table{\figpath ssr_NNBPDN_gauss_s_5_snr_40.dat};

\addplot [
color=red,
thick,
densely dashdotted,
mark=*,
mark options=solid,
mark size=1pt
]
table{\figpath ssr_NNLS_gauss_s_5_snr_40.dat};

\makeatletter

\end{groupplot}
\end{tikzpicture}
  \caption{Comparison of the two non-negative data models for an input SNR of $40~\dB$, $\sparsity = 5$, $\csmeasdim = 50$ and $\cssignaldim = 200$. \emph{left:} MSE on sparse support $\supp$. \emph{right:} sparse support recovery by thresholding. Note the logarithmic scaling of the \emph{left} ordinate.}
  \label{fig:mse_sup_ssr_combi_uni_vs_gauss}
\end{figure}
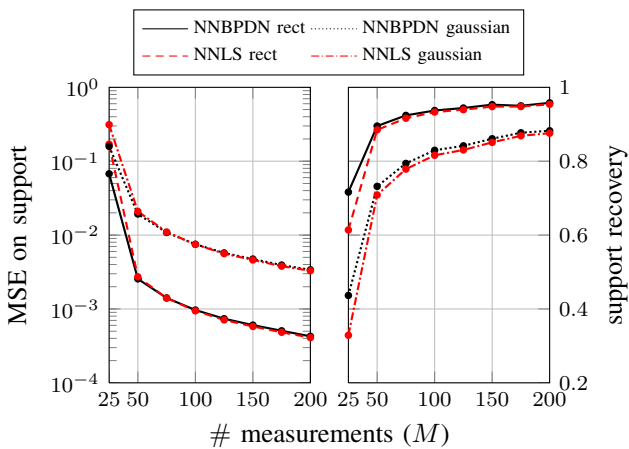

So far, all results were presented for the uniform (rect) data model. A comparison between the recovery performance of the uniform and the Gaussian data model is shown in \ref{fig:mse_sup_ssr_combi_uni_vs_gauss}. Similar to the results in \ref{sec:olf_eval}, the Gaussian data model generally performs worse. The general effects discussed above are similarly present with the Gaussian data model, however.

Summarizing, we see that the recovery performance of the discontinuous network is very similar to the performance obtained by solving a NNBPDN optimization problem. The NNBPDN approach tends to recover the sparse support slightly better. For most choices of the parameters, the dynamic system features a marginally lower MSE on the support, this difference becomes more profound in low SNR situations. Nevertheless, we expect these differences to be smaller in practice since the regularization parameter $\csreg$ of the NNBPDN was chosen as the best from a set of $50$ values for each Monte Carlo instance, which cannot be done in practice. Thus, the discontinuous network shows a very promising recovery performance for non-negative sparse approximation applications.

\subsection{Relation to existing work}
As briefly summarized in \ref{sec:introduction}, several networks exist which are used in similar applications. Firstly, the most profound novelty of our presented system \ref{eq:disc_sys} is that the integrators are discontinuous and have a lower limit equal to zero (or are individually upper- and lower-bounded, compare \ref{sec:box_constraints}). In contrast to this, in related systems (cf. \cite{bouzerdoum_neural_1993,xia_new_1996,rozell_sparse_2008}), the output of the integrators is not constrained. Although for some of these systems, the output is directly processed by a nonlinear thresholding function before being connected to neighboring nodes, all of these systems still feature unbounded internal system states. This also influences the convergence behavior of said systems. The limited integrator in our system captures inherent limits, e.g., on neuronal firing rates, directly. However, careful stability analysis is required due to the discontinuous nature of the system. Secondly, our original application for which our neural network was investigated (see \ref{sec:olf_model}) features self-inhibition that violates the necessary conditions on stability in classical continuous HNNs.

\section{Conclusion}
\label{sec:conclusion}
In this work, we investigated a discontinuous neural network that is applied as a model of the mammalian olfactory system and more generally as a method for solving non-negative sparse approximation problems. The discontinuities in the system function of the neural network arise from the fact that the integrators of the system are bounded to having non-negative outputs. Hence, the integrators ``switch'' between being active if the output is greater than zero and being inactive if the output reaches zero.
First of all, we showed that the presented neural networks features equilibrium points that correspond to the solutions of a general non-negative least squares problem. We then specified a Caratheodory solution to the system's discontinuous differential equations and proceeded to prove that the network is stable, provided that the system matrix has full column-rank, using Lyapunov's direct method. Moreover, we were able to show, under a mild condition on the equilibrium point, that the system performs only a finite amount of switches while converging. We then generalized the obtained results to the case where the integrators of the system have arbitrary lower- and upper limits. In this case, the neural network converges to the solution of a bounded least squares optimization problem. 
Two applications for the presented neural network were shown. Firstly, our original application was shown, where the neural network was derived from modeling of the mammalian olfactory system. We used this model to investigate a potential application of the olfactory system, which is the decomposition of a mixture of chemical stimuli from noisy neural responses. Based on our findings, we argued that the olfactory system may be capable of performing such a complex sparse recovery task. Secondly, we generalized the application to embrace non-negative sparse approximation problems. We showed that the stability of the neural networks is unaffected under this model, even if the system matrix does not possess full column-rank. Then, we compared the recovery performance of the network with a classical non-negative basis pursuit denoising algorithm. Since the recovery performance differs only marginally, the presented network can be successfully applied. Compared to the classical algorithm, the network is independent from the choice of a regularization parameter, which strongly affects the recovery performance of the classical non-negative basis pursuit denoising algorithm.

\appendix
\section*{Input SNR Derivation}
Let $\imat_\supp$ denote a sparse identity matrix, i.e., a diagonal matrix with ones only at positions $\{(i,i)~|~i \in \supp\}$ and zeros otherwise. Likewise, let $\onevec_\supp$ denote a sparse one-vector containing ones solely at indices $\{i~|~i\in \supp \}$ and zeros otherwise.
For the derivation of the input signal power $\expop{\transp{\vec{\olfin}_\olftrue}\vec{\olfin}^{}_\olftrue}$, we need $\expop{\transp{\mat{\olfmat}}\mat{\olfmat}}$, $\traceop{\expop{\transp{\mat{\olfmat}}\mat{\olfmat}} \imat_\supp}$ and $\transp{\onevec_\supp} \, \expop{\transp{\mat{\olfmat}}\mat{\olfmat}} \onevec^{}_\supp$.
Considering the independencies between the entries we get:
\begin{align*}
\expop{\transp{\mat{\olfmat}}\mat{\olfmat}}_{ij} = \expop{\sum_{k = 1}^{\olfmatrowdim} \matidx{\olfmat}{k}{i} \, \matidx{\olfmat}{k}{j}} = \olfmatrowdim \, \expop{\matidx{\olfmat}{1}{i}\,\matidx{\olfmat}{1}{j}} \\
\Rightarrow \expop{\transp{\mat{\olfmat}}\mat{\olfmat}} = \frac{\matmean^2}{\matmean^2 + \matvar} \, \onevec \transp{\onevec} + \frac{\matvar}{\matmean^2 + \matvar} \imat \,.
\end{align*}
Based on this result it follows that
\begin{equation*}
\traceop{\expop{\transp{\mat{\olfmat}}\mat{\olfmat}} \imat_\supp} = s \,,
\end{equation*}
and
\begin{align*}
\transp{\onevec_\supp} \, \expop{\transp{\mat{\olfmat}}\mat{\olfmat}} \onevec^{}_\supp &= \frac{\matmean^2}{\matmean^2 + \matvar} \, \transp{\onevec_\supp} \onevec \transp{\onevec} \onevec_\supp + \frac{\matvar}{\matmean^2 + \matvar} \transp{\onevec_\supp} \onevec^{}_\supp \\
&= \frac{\sparsity^2 \matmean^2 + \sparsity \, \matvar}{\matmean^2 + \matvar} \,.
\end{align*}
Finally, we can calculate the input signal power $\expop{\transp{\vec{\olfin}_\olftrue}\vec{\olfin}^{}_\olftrue}$ using the results obtained above:
\begin{align*}
\expop{\transp{\vec{\olfin}_\olftrue}\vec{\olfin}^{}_\olftrue} &= \expop{\transp{\vec{\olfmcout}_\olftrue}\transp{\mat{\olfmat}}\mat{\olfmat}\vec{\olfmcout}^{}_\olftrue} = \expop{\traceop{\transp{\mat{\olfmat}}\mat{\olfmat} \, \vec{\olfmcout}^{}_\olftrue\transp{\vec{\olfmcout}_\olftrue}}} \\
&= \traceop{\expop{\transp{\mat{\olfmat}}\mat{\olfmat}} \expop{\vec{\olfmcout}_\olftrue\transp{\vec{\olfmcout}_\olftrue}}} \\
&= \traceop{\expop{\transp{\mat{\olfmat}}\mat{\olfmat}} (\covop{\vec{\olfmcout}_\olftrue} + \expop{\vec{\olfmcout}_\olftrue}\transp{\expop{\vec{\olfmcout}_\olftrue}})} \\
&= \traceop{\expop{\transp{\mat{\olfmat}}\mat{\olfmat}} (\truemixvar \, \imat_\supp + \truemixmean^2 \, \onevec^{}_\supp\transp{\onevec_\supp})} \\
&= \truemixvar \, \traceop{\expop{\transp{\mat{\olfmat}}\mat{\olfmat}} \imat_\supp} +  \truemixmean^2 \, \transp{\onevec_\supp} \, \expop{\transp{\mat{\olfmat}}\mat{\olfmat}} \onevec^{}_\supp \\
&= \truemixvar \sparsity + \frac{\truemixmean^2 (\sparsity^2 \matmean^2 + \sparsity \, \matvar)}{\matmean^2 + \matvar} \,.
\end{align*}
The noise power is simply $\expop{\transp{\vec{\olfnoise}}\vec{\olfnoise}} = \olfmatrowdim \, \noisevar$. Thus, the input SNR follows as \ref{eq:input_snr}.


\ifCLASSOPTIONcaptionsoff
  \newpage
\fi



\bibliographystyle{IEEEtran}
\bibliography{literature}
\end{document}